\documentclass[mnsc,nonblindrev]{informs3} 

\DoubleSpacedXI

\usepackage{natbib}
 \bibpunct[, ]{(}{)}{,}{a}{}{,}%
 \def\bibfont{\small}%
\usepackage{hyperref}
\usepackage{xcolor}
\hypersetup{
	colorlinks,
	linkcolor={red!50!black},
	citecolor={blue!50!black},
	urlcolor={blue!80!black}
}
\usepackage{float}
\usepackage{algorithm}
\usepackage[noend]{algpseudocode}
\usepackage{framed} 
\usepackage{xcolor} 
\usepackage{graphicx}
\usepackage{epstopdf}
\usepackage{subfigure}
\graphicspath{ {images/} }
\usepackage{booktabs}
\usepackage{makecell} 
\usepackage{multirow}
\usepackage{bbm}
\usepackage{threeparttable} 
\usepackage{enumerate}
\usepackage{xspace}

\newcommand{\nSamples}{n}

\newcommand{\nLayers}{L} 
\newcommand{\layerN}{l}
\newcommand{\LReNVC}{LR-$\varepsilon$NVC\xspace}
\newcommand{\LReNVCR}{LR-$\varepsilon$NVC-R\xspace}
\newcommand{\LRNVC}{LR-NVC\xspace}
\newcommand{\LRMSE}{LR-MSE\xspace}
\newcommand{\NNeNVC}{NN-$\varepsilon$NVC\xspace}
\newcommand{\NNNVC}{NN-NVC\xspace}
\newcommand{\NNMSE}{NN-MSE\xspace}
\newcommand{\inNVLoss}{\mathcal{L}^{\varepsilon NV}\xspace}

\newcommand{\LReNVCa}{LR-\varepsilon NVC\xspace} 
\newcommand{\LReNVCRa}{LR-\varepsilon NVC-R\xspace} 
\newcommand{\LRNVCa}{LR-{NVC}\xspace}
\newcommand{\LRMSEa}{LR-{MSE}\xspace}
\newcommand{\NNeNVCa}{NN-{\varepsilon NVC}\xspace}
\newcommand{\NNNVCa}{NN-{NVC}\xspace}
\newcommand{\NNMSEa}{NN-{MSE}\xspace}
\TheoremsNumberedThrough     
\ECRepeatTheorems

\EquationsNumberedThrough    

\begin{document}


\RUNAUTHOR{Chen, Fu, Huang, and Bai}

\RUNTITLE{Custom $\varepsilon$-insensitive Operational Costs for Censored Observations}

\TITLE{Learning Decisions Offline from Censored Observations with $\varepsilon$-insensitive Operational Costs}

\ARTICLEAUTHORS{%
\AUTHOR{Minxia Chen}
\AFF{Lingnan College, Sun Yat-sen University, Guangzhou, China 510275, \EMAIL{chenmx75@mail2.sysu.edu.cn}} 
\AUTHOR{Ke Fu, Teng Huang\footnote{Corresponding author}}
\AFF{School of Business, Sun Yat-sen University, Guangzhou, China 510275, \EMAIL{\{fuke, huangt258\}@mail.sysu.edu.cn}}
\AUTHOR{Miao Bai}
\AFF{Department of Operations and Information Management, School of Business, University of Connecticut, Storrs, U.S. 06269,
\EMAIL{miao.bai@uconn.edu}}
} 

\ABSTRACT{
Many important managerial decisions are made based on censored observations.
Making decisions without adequately handling the censoring leads to inferior outcomes.
We investigate the data-driven decision-making problem with an offline dataset containing the feature data and the censored historical data of the variable of interest without the censoring indicators. 
Without assuming the underlying distribution, we design and leverage $\varepsilon$-insensitive operational costs to deal with the unobserved censoring in an offline data-driven fashion.
We demonstrate the customization of the $\varepsilon$-insensitive operational costs for a newsvendor problem and use such costs to train two representative ML models, including \textit{linear regression} (LR) models and \textit{neural networks} (NNs).
We derive tight generalization bounds for the custom LR model without regularization (\LReNVC) and with regularization (\LReNVCR), and a high-probability generalization bound for the custom NN (\NNeNVC) trained by \textit{stochastic gradient descent}.
The theoretical results reveal the stability and learnability of \LReNVC, \LReNVCR ~and \NNeNVC.
We conduct extensive numerical experiments to compare \LReNVCR ~and \NNeNVC ~with two existing approaches, \textit{estimate-as-solution} (EAS) and \textit{integrated estimation and optimization} (IEO).
The results show that \LReNVCR ~and \NNeNVC ~outperform both EAS and IEO, with maximum cost savings up to 14.40\% and 12.21\% compared to the lowest cost generated by the two existing approaches.
In addition, \LReNVCR's and \NNeNVC's order quantities are statistically significantly closer to the optimal solutions should the underlying distribution be known.
The custom learning algorithms with $\varepsilon$-insensitive operational costs can effectively learn decisions from censored observations and reduce decision error.
}%





\maketitle

%


\section{Introduction}
\label{sec:intro}

In many environments, decision-makers can only observe imperfect values of the variables of interest.
For example, in lost-sales inventory systems, retailers cannot observe the lost sales. They have to make decisions such as pricing \citep{bu2022offline} and inventory replenishment \citep{ban2020confidence} based on sales, i.e., the censored demand observations.  
Censored observations might lead to inferior decisions. 
Individuals tend to rely heavily on the observed censored sample, biasing their belief about the underlying population. Censorship bias exacerbates higher degrees of censorship, higher variance in the population, and higher variability in the censorship points \citep{feiler2013biased}. 
When the decision-maker does not know whether a censoring occurs, which often is the most realistic scenario \citep{ban2020confidence}, the decision problem becomes significantly more challenging.
We attempt to tackle this challenge by proposing a set of one-step offline learning algorithms with $\varepsilon$--insensitive cost functions.  

\subsection{Decision-Making with Unobserved Censoring}
\label{sec:examples}
In various contexts, decision-makers often have only censored historical observations of key variables, but do not know whether they are censored or not.
Here, we provide three exemplary problems. 
They each represent a large body of fundamental decision-making problems in operations management.

\subsubsection{Newsvendor with Sales Data.}

A newsvendor faces stochastic demand in each period.
She needs to determine the order quantity prior to the demand realization given past \textit{sales} and feature values related to the demand.  
The unmet demand is lost, and the leftover inventory is disposed of at the end of the period. 
The ordering objective is to minimize the total expected cost, including the underage and overage costs due to stockout and overstock, respectively.
When only the sales data is available and the historical inventory information is not, demand censoring might have happened, but the newsvendor does not know.
\cite{shmueli2005useful} utilize a real-world dataset consisting of the quarterly sales of a well-known brand of a particular article of clothing at the stores of a large national retailer without historical inventory information.  
Also, the real-world dataset in \citet{oroojlooyjadid2020applying} consists of basket data from a retailer with only sales data without inventory information.

\subsubsection{Custom Pricing.}

Sellers of unique products or services face the challenge of setting the price that would maximize their revenue.
Ideally, the seller should price at the buyer's reservation price. However, this is generally not possible, since the seller does not know the buyer's reservation price.
In fact, if the price is higher than the buyer's reservation price, the customer will not accept the offer, and the seller ends up earning nothing.
On the other hand, if the price is lower than the reservation price, the buyer will accept the offer, but obviously, the seller will suffer a loss compared with the ideal pricing.
\cite{braden1991informational} describe a problem of pricing a unique durable good (e.g., a house).
The seller cannot observe the customer's reservation price or willingness to pay.
However, the seller can be certain that the customer's reservation price is higher than or equal to the accepted price.
In other words, the accepted price is a censored observation of the reservation price.
\citet{ye2018customized} describe a similar pricing problem for room listings on Airbnb. Each listing offers a ``unique value and experience.''
The authors design algorithms to determine the price of the listings so that the rooms can be booked at a higher price.

\subsubsection{Preventive Replacement.}

In a preventive replacement problem \citep{fox1967adaptive}, the decision-maker sets a time interval, and a piece of equipment is replaced at the time of failure or at the end of the planned time interval.
The cost of a failure replacement is much more than that of a planned replacement.
Nonetheless, it is not economical to schedule replacement frequently.
Therefore, the objective is to determine the replacement interval to minimize the total expected cost.
In these cases, the decision-maker can only observe the failure time of the equipment when it actually fails; other replacements are at scheduled replacement intervals.
If the replacement interval is set too small, there would be no failure observed but a history of scheduled replacements.
In this case, all preventive replacement intervals are censored observations of the actual failure times.

\subsection{Model Overview}
Denote by $y \in \mathcal{Y} \subseteq \mathbb{R}^{dim_y}$ the decision, by $\{d_1, \ldots, d_n\}$ the data on the uncertain quantities $D \in \mathcal{D} \subseteq \mathbb{R}^{dim_d}$ that are closely related to the decision, and by $\{\boldsymbol{x}_1, \ldots, \boldsymbol{x}_n\}$ the data on the covariate $\boldsymbol{X} \in \mathcal{X} \subseteq \mathbb{R}^{dim_x}$. 
One searches for the optimal decision that minimizes the expectation of cost $c(\cdot)$. 
The focus of this study is scenarios when the observations of $D$ are censored and the censoring indicators are unavailable to the decision-maker.
We formulate the aforementioned type of data-driven decision-making problem as
\begin{equation}
    \min_{y \in \mathcal{Y}} \mathbb{E}_D [c(y; D | \boldsymbol{X} = \boldsymbol{x})]. \label{eq:bertsimas}
\end{equation}

The problem in Equation \eqref{eq:bertsimas} takes the expectation over the random variable $D$. 
If the (feature-dependent) distribution of $D$ is known, the above problem can be solved and a distribution-related solution is obtained. 
For example, in the classical newsvendor problem, the solution of Equation \eqref{eq:bertsimas} is the quantile of the demand distribution (see, e.g., \cite{zipkin2000foundations}). 
However, in reality, the distribution of $D$ is unknown, and only historical observations of $D$ and related feature data are available. In the data-driven method, a two-step approach is to first estimate $D$ based on the historical data, and then solve the optimization problem in Equation \eqref{eq:bertsimas}. However, this kind of approach would introduce prediction errors into the optimization problem, leading to suboptimal solutions \citep{ban2019big}. To this end, some propose one-step methods, integrating estimation and optimization, to obtain decisions directly from data. Our proposed algorithms belong to this kind of one-step method. 

In the stream of the one-step algorithm, to estimate the expectation in Equation \eqref{eq:bertsimas}, \cite{bertsimas2020predictive} propose a framework and related methods for selecting samples and setting weights. Their estimations are in the form of the following equation:
\begin{equation}
    \min_{y \in \mathcal{Y}} \sum_{i=1}^n w_{n,i}(\boldsymbol{x}) c(y;d_i),\label{eq:ber_sol}
\end{equation}
where $w_{n, i}(\boldsymbol{x})$ is the weight derived from sample data.
In line with another stream of studies \citep{sachs2014data,ban2019big,huber2019data,oroojlooyjadid2020applying}, we utilize the \textit{empirical risk minimization} (ERM) principle; therefore, our objective can be written as
\begin{align}
    & \min \frac{1}{n} \sum_{i=1}^{n} c(y_i; d_i), \label{eq:erm1}
\end{align}
where $c(\cdot)$ is the operational cost of making decision $y_i$ for the $i$th observation $d_i$.
For example, when determining the newsvendor order quantities, 
\cite{ban2019big,huber2019data}, and \cite{oroojlooyjadid2020applying} use the newsvendor cost as $c(\cdot)$ and propose data-driven-solution methods.
These papers assume that demand is perfectly observed.
However, when the demand in the newsvendor problem is censored, and more generally, when the observations of key variables for the decision-making are not fully observed or accurately recorded, the operational cost computed using censored data does not reflect the actual cost.
In this study, we aim to solve the problem in Equation~\eqref{eq:erm1} when $d_i$ is censored and the information of whether the observation is censored or not is not available to the decision-maker.

Our algorithms allow the decision-maker to specify the cost function $c(\cdot)$ that reflects the operational cost and further customize this operational cost to deal with the unobserved censored observations of key variables for the decision.
We take advantage of  $\varepsilon$-insensitive loss functions for handling censored observations.
The $\varepsilon$-insensitive loss function is first used in \textit{support vector} (SV) methods \citep{vapnik1999nature}.
It allows some deviations for a few anomalous data points' predictions.
Hence, SV methods have shown a great performance advantage over traditional methods \citep{smola2004SVR,ho2012large} and have good generalization performance \citep{zhang2020predictive}.
In this study, we design the $\varepsilon$-insensitive operational cost to guide the decision prescription, considering that part of the observations may be censored.

Our proposed algorithms directly prescribe decision $y_i$ from the censored value $s_i$ for $d_i$ and its related features $\boldsymbol{x}_i$.
We denote by $\boldsymbol{\theta}$ the parameter vector that specifies the mapping from features to the decision.
The decision problem is finalized as 
\begin{align}
    & \min \frac{1}{n} \sum_{i=1}^{n} c(y_i(\boldsymbol{x}_i;\boldsymbol{\theta}); s_i). \label{eq:erm2}
\end{align}
We utilize \textit{stochastic gradient descent} (SGD) \citep{robbins1951sgd} to minimize the predefined operational cost and obtain $\boldsymbol{\theta}$'s estimation.

\subsection{Summary of Results and Contributions}

We propose an offline decision-learning framework that directly prescribes decisions based on the censored offline dataset via one-step machine-learning algorithms.
Specifically, we demonstrate the framework with a multi-feature newsvendor problem. 
We choose this setting because the newsvendor is a foundational problem in operations management and many other problems have a similar informational structure \citep{braden1991informational}. 
We show how to specify the cost function that reflects the operational cost and further customize this operational cost to deal with the unobserved censored observations of key variables for the decision. We take advantage of $\varepsilon$-insensitive loss functions for handling censored observations. 
We then showcase how to integrate the proposed loss functions with two popular \textit{machine learning} (ML) methods, \textit{linear regression} (LR) and \textit{neural networks} (NNs).
The three proposed algorithms are named \LReNVC, \LReNVCR ~(with regularization) ~and \NNeNVC.

We derive the stability properties and generalization bounds for \LReNVC, \LReNVCR ~and \NNeNVC ~in Section~\ref{sec:prop}.
\LReNVC ~is a deterministic learning algorithm and can be formulated as a \textit{linear programming} (LP) problem.
It shows strong stability with the stability parameter scaled as the inverse of the sample size $\frac{1}{n}$ (Proposition~\ref{prop:LR-stable}).
\LReNVC ~also has the tight generalization bound scaled as $O(p/\sqrt{n})$, where $p$ is the number of features (Theorem~\ref{theorem:LR-GenBound}).
\LReNVCR ~shows strong stability with the stability parameter scaled as $O(p/(n\lambda))$, where $\lambda$ is the regularization parameter (Proposition~\ref{prop:LR-R-stable}). Its tight generalization bound is scaled as $O(p/(\sqrt{n}\lambda))$ (Theorem~\ref{theorem:LR-R-GenBound}).
For \NNeNVC ~trained by $K$-pass SGD, a randomized learning algorithm, we derive the bound on its \textit{uniform argument stability} (UAS, \citealt{bassily2020stability}), which is scaled as $\frac{1}{\sqrt{n}}$ or $\frac{1}{K}$ with proper learning rates (Proposition~\ref{prop:DNN-UASBound}). 
Then, we derive \NNeNVC's high probability generalization bound (Theorem~\ref{theorem:dnnGBound}).

We numerically demonstrate the effectiveness of our proposed methods concerning the out-of-sample performance.
We compare our proposed algorithms with two baseline approaches, the \textit{Estimate-as-Solution} (EAS) approach \citep{oroojlooyjadid2020applying} and the \textit{Integrated Estimation and Optimization} (IEO) approach \citep{ban2019big}. 
The numerical results highlight the value of our proposed algorithm in better dealing with censoring.
In terms of the out-of-sample empirical operational cost, we first find that IEO's average newsvendor cost on test sets is lower than that of EAS methods, consistent with the results of \citep{oroojlooyjadid2020applying} for uncensored cases.
In addition, at different censoring levels, \LReNVCR's average newsvendor cost on test sets is 2.45\% to 14.40\% less than that of LR-based IEO, and \NNeNVC's average newsvendor cost on test sets is 3.39\% to 12.21\% less than that of NN-based IEO.
The cost-saving percentages increase as the censoring gets severe.
In order quantities, the ones prescribed by \LReNVCR ~and \NNeNVC ~are statistically significantly closer to the optimal decisions should the underlying demand distribution be known to the decision-maker.

In the following, we review related work in Section~\ref{sec:litreview}.
We describe the one-step offline decision-learning framework with censored observations and discuss its theoretical properties in Section~\ref{sec:framework}.
We conduct extensive numerical experiments with our proposed algorithms for a repeated newsvendor problem and discuss the results in Section~\ref{sec:numExp}.
Lastly, we conclude in Section~\ref{sec:conclusion}.

\section{Literature Review}
\label{sec:litreview}

Our work relates to learning for decision-making, business decision-making based on censored observations, and the design and application of $\varepsilon$-insensitive loss functions. 
In the following, we elaborate on each of these research streams.

\subsection{Data-Driven Decision-Making under Uncertainty}
\label{subsec:review Data-driven}

 We take a data-driven approach to prescribing decisions based on offline data.
One approach to incorporating rich data into decision-making processes is to utilize point estimations generated by ML algorithms in formulating the decision problems. For example, \cite{ferreira2016analytics} estimate demand from sales and then insert the estimations into the pricing problem.

A second approach is to embed entire ML model structures in the decision problems' formulation, such as the objective function.
For example, \cite{huang2019predictive} embed LR and \textit{support vector regression} (SVR) models with radial and linear kernels into a facility location problem. 
The decision variables consist of the ML algorithms' inputs. 
The ML algorithms' actual outputs depend on the decision variables' values.
\cite{bergman2022janos} further develop a solver wrapper that supports the embedding of LR models, \textit{logistic regression} models, and NNs into the objective function of a user-defined optimization problem.
In this group of studies, the predictions of the ML algorithms are realized when the optimization problem is solved.
Also, the solutions to the ensuing optimization problem are the decisions we seek. 
This is one of the key differences from the research stream we summarize next, in which decisions are the outputs of the ML process.

A third approach is called single-step machine-learning algorithms \citep{ban2019big}, also known as the end-to-end prediction and optimization approach \citep{qi2023practical,honguyen2022risk}. These approaches leverage ML techniques and get actionable decisions directly from the output of the ML process. It is achieved by minimizing the context-related operational costs while training the ML algorithms. For example, one of the proposed methods in \cite{ban2019big} assumes a linear relationship between the ordering decision and the related features, and embeds newsvendor cost functions into linear regression models. By minimizing the newsvendor cost, this approach estimates the parameters in the LR model and obtains the optimal order quantity.
\cite{oroojlooyjadid2020applying} take advantage of the NN structure and SGD algorithm that directly minimize the newsvendor cost to obtain newsvendor order quantities. 
Although, neither of these two papers takes into account the fact that sales data are censored observations of demand.

In line with the third stream, we propose an offline decision-learning framework that handles censored observations via one-step machine-learning algorithms. Specifically, we customize the operational costs so that the algorithms are able to produce decisions that can recover the censored observations to some extent.

\subsection{Censored Observations in Business Decision-Making}

Prior literature has documented various approaches to handling censored data when knowing whether the censoring happens, such as survival analysis \citep{camuffo2020scientific,musalem2021retail,rivera2021traditional} and Bayesian approaches \citep{besbes2022exploration}.
Survival analysis applies statistical methods for time-to-event data, characterized by the survival probability function and the hazard function. 
Bayesian approaches assume the random variable follows a distribution characterized by unknown parameters. The unknown parameters are dynamically learned from prior distribution assumptions. 
In this paper, we propose learning algorithms for decision-making using censored data that do not require knowledge of the problem's underlying distribution.
In particular, we study an offline learning problem with censored historical data. 

Despite offline datasets with censored data being widely encountered in various decision-making contexts, fewer than a handful of published papers in operations research and management science study offline learning with censored information.
\cite{sachs2014data} use point-of-sales data to establish sales patterns and use the patterns to estimate the lost sales in the data-driven newsvendor problem. 
\cite{huber2019data} also use point-of-sales data to estimate demand in the numerical study.
\cite{ferreira2016analytics} use a similar approach to estimating demand from sales data for a pricing problem.
\cite{bertsimas2020predictive} handle the censored observations in the numerical experiment by developing a conditional variant of the Kaplan-Meier method \citep{kaplan1958nonparam} for the weights in the weighted empirical risk (i.e., Equation~\eqref{eq:ber_sol}). 
\cite{bu2022offline} define an ambiguity set of distributions to capture the uncertainty created by the information loss in observed censored data in a single pricing problem with an offline dataset of historical price, inventory level, and potentially censored sales.
We develop an offline learning framework to learn the relationship between exogenous features and the decision, allowing for the correction of the censored target values during the learning process.

Although there is a rich literature on decision-making with censored observations, the existing research focuses mostly on observed censoring. That is, the decision-maker has knowledge of whether the observations are censored or not. There is rather limited research considering unobserved censoring, in which case the decision-maker does not know whether an observation is censored. 
One exception is \cite{ban2020confidence}. She studies a multiperiod inventory system and proposes a nonparametric estimation procedure in an offline setting using censored demand to obtain confidence intervals for the $(S, s)$ policy. 
For the unobserved censoring case, she provides two specific conditions for the relative position of past stocking levels to the optimal order-up-to levels, following which the unobserved censoring case can be reduced into the uncensored or observed censored case. However, the decision-maker cannot know a priori whether either condition is satisfied or not.
Unlike \cite{ban2020confidence}, we develop an offline learning framework for a general data-driven problem with unobserved censoring. We leverage custom ML methods to handle unobserved censored observations in decision-making. 

In the meantime, there are numerous studies in online learning settings (e.g., \citealp{chen2020dynamic,chen2020data,zhang2020closing,chen2021data,yuan2021marrying}), in which censored observations are related to the decision-maker's actions and decisions affect future data. Nevertheless, they are not the focus of this paper.

\subsection{$\varepsilon$-insensitive Loss Functions}

Our proposed $\varepsilon$-insensitive operational cost is inspired by the loss function used in \textit{support vector machines} (SVMs) \citep{vapnik1999nature}. 
SVMs have achieved enormous success in classification problems \citep{vapnik1999nature}, regression (\citealp{smola2004SVR,ho2012large}), clustering \citep{lee2005improved}, ranking \citep{chu2007support}, etc. \cite{zhang2020predictive} uncover an SVR-based approach's strength in dealing with missing feature data from data providers that strategically conceal information to gain a favorable outcome.

Among these successes, \cite{shivaswamy2007support} and \cite{khan2008support} extend SVMs to the case of censored targets in survival analysis by customizing the $\varepsilon$-insensitive objective function in the SV algorithm.  
\cite{ye2018customized} use a custom $\varepsilon$-insensitive loss function to prescribe dynamic pricing decisions while accounting for the fact that the optimal price is greater than or equal to the list price when a listing is booked. In contrast, when a listing is not booked, the optimal price is less than or equal to the list price.

This study proposes customizing the operational cost to obtain decisions from censored values. The custom operational costs leverage the $\varepsilon$-insensitive structure.

\section{Offline Decision-Learning Algorithms with Censored Observations}
\label{sec:framework}

This section presents a framework that prescribes decisions offline based on censored historical data.
Section~\ref{sec:three_examples} describes how to customize $\varepsilon$-insensitive operational cost functions to handle censored values. 
Then, in Section~\ref{sec:CustML}, we demonstrate how to integrate the customize loss functions into existing ML algorithms. 
Lastly, we show the properties of the decision-learning process in Section~\ref{sec:prop}.

\subsection{Custom $\varepsilon$-insensitive Operational Costs}
\label{sec:three_examples}

This study uses $\varepsilon$-insensitive loss functions to handle the censoring.
The algorithm twists the censored value in the direction of the actual value.
As censoring is prevalent in decision-making, such an idea could be relevant in many contexts.
We demonstrate how to customize the operational cost leveraging the $\varepsilon$-insensitive structure using the three illustrative examples introduced in Section ~\ref{sec:examples}.

\subsubsection{Newsvendor Ordering.}

Denote the unit overage cost by $c_o$ and the unit underage cost by $c_u$.
Denote by $d_i$ the true demand of the $i$th period, by $s_i$ the observed sales, and by $y_i$ the order quantity.
In the literature on data-driven newsvendor problems with perfect demand observations (e.g., \citealt{beutel2012safety,ban2019big,huber2019data,oroojlooyjadid2020applying}), 
the newsvendor incurs the overage cost when $y_i>d_i$ and the underage cost when $y_i<d_i$,
and the loss for the solution algorithm to minimize is given by
\begin{equation}
    \mathcal{L}^{NV} = \frac{1}{n}\sum_{i=1}^{n} \left[c_o(y_i-d_i)^+ + c_u(d_i-y_i)^+ \right]. \label{eq:nvc-uncensor}
\end{equation}

In the focal context, $d_i$ is unobserved, and only $s_i$, the satisfied demand, is recorded.
We define the $\varepsilon$-insensitive newsvendor cost loss as follows:
\begin{align}
    \tilde{\mathcal{L}}^{\varepsilon NV} &= \frac{1}{n}\sum_{i=1}^{n} \left[c_o(y_i-s_i-\varepsilon_1)^+ + c_u(s_i-y_i+\varepsilon_2)^+ \right], & \varepsilon_1 > \varepsilon_2 \geq 0. \label{eq:nvc-censored}
\end{align}
Following Equation~\eqref{eq:nvc-censored}, the learning algorithm increases the newsvendor's overage cost when she orders more than $s_i + \varepsilon_1$. This is how the learning algorithm considers the fact that $s_i$ might be censored and, as such, encourages ordering more than $s_i$.
In the meantime, the learning algorithm accumulates the newsvendor's underage cost if she orders less than $s_i + \varepsilon_2$.
The learning algorithm will ignore costs when the order quantity $y_i$ falls in the range $[s_i+\varepsilon_2,s_i+\varepsilon_1]$.
In other words, the learning algorithms consider order quantities in the range $[s_i+\varepsilon_2,s_i+\varepsilon_1]$ perfect by assigning no loss for them.
$\varepsilon_1$ and $\varepsilon_2$ are two hyperparameters of the learning algorithm, which would be tuned using cross-validation method during model training.
Lastly, we normalize the loss values by defining $\alpha = \frac{c_u}{c_u+c_o}$ and transform Equation~\eqref{eq:nvc-censored} into the following formula:
\begin{align}
    \mathcal{L}^{\varepsilon NV} &= \frac{1}{n}\sum_{i=1}^n \left[(1-\alpha) (y_i - s_i - \varepsilon_1)^+ + \alpha (s_i + \varepsilon_2 - y_i)^+ \right],  & \varepsilon_1 > \varepsilon_2 \geq 0. \label{eq:le1e2}
\end{align}

\subsubsection{Custom Pricing.}

Let $d_i$ be the reservation price of the $i$th customer and $s_i$ be the selling price.
The seller observes $d_i$ only when $d_i\geq s_i$ because when $d_i < s_i$, the customer turns down the offer.
The problem is to decide the optimal price $y_i$ based on past observations $s_i$.
We define the $\varepsilon$-insensitive cost function as 
\begin{align}
    \tilde{\mathcal{L}}^{\varepsilon CP} &= \frac{1}{n} \sum_{i=1}^n \left[
    c_2 (s_i - y_i)^+ + c_1 (y_i - s_i - \varepsilon)^+ \right], & \varepsilon>0,
    \label{eq:cp}
\end{align}
where $\varepsilon>0$ is a hyperparameter in the algorithm.
The decision-maker suffers a larger unit loss of $c_2$ if the price $y_i$ is set lower than the observed price $s_i$; no loss if $y_i$ is in the range of $[s_i, s_i + \varepsilon]$; and a smaller unit loss of $c_1$ if $y_i$ is higher than $s_i + \varepsilon$.

\subsubsection{Preventive Replacement.}

Let $c_1$ denote the cost of a planned replacement and $c_2$ the cost of a failure replacement, where  $c_2>c_1>0$.
Let $s_i$ be the historically observed replacement time; $d_i$ be the \textit{actual} failure time, which is unknown; and $y_i$ be the ideal length of the replacement interval.
We modify the cost function in \cite{fox1967adaptive} and present the custom operational cost for preventive replacement as follows:
\begin{align}
    \tilde{\mathcal{L}}^{\varepsilon RP} &= \frac{1}{n} \sum_{i=1}^n \left[ c_2 e^{-s_i} \mathbb{I}(y_i - s_i - \varepsilon > 0)  + c_1 e^{-s_i} \mathbb{I}(s_i + \varepsilon - y_i > 0)  \right], & \varepsilon>0. \label{eq:pr}
\end{align}
where $y_i$ is the length of the replacement interval to be set; $\varepsilon>0$ is a hyperparameter in the algorithm, the value of which will be determined during the training; and $s_i$ comes from the data.

Specifying Equation~\eqref{eq:pr} as the cost means that, when $y_i > s_i + \varepsilon$, i.e., the scheduled replacement time is longer than the observed replacement time, a high unit cost $c_2$ is incurred for failure replacement; when $y_i < s_i + \varepsilon$, i.e., the scheduled replacement time is sooner than the observed replacement time, a low unit cost $c_1$ is incurred for the planned replacement.

\subsection{Custom Machine Learning Algorithms}
\label{sec:CustML}

This section introduces how to integrate the proposed loss functions with two frequently used ML models, LR and NNs, taking the newsvendor problem as an example.
We denote the two new custom algorithms by \LReNVC, \LReNVCR ~(with regularization) ~and \NNeNVC.
The decision-maker collects $n$ observations. Each observation contains a $p$-dimentional feature vector $\boldsymbol{x}_i$ and an unobserved censored value $s_i$ that relates to the cost of the decision.
We denote by $d_i$ the unknown true value, by $y_i$ the output decision of the framework, and by $S_n = \{\boldsymbol{x}_i,s_i\}_{i=1}^n$ the offline dataset.
We assume that the elements of $S_n$ are \textit{independent and identically distributed (i.i.d.)} samples from an unknown distribution. 
We make the assumption for the feature vector $\boldsymbol{x}$ that $x_1=1$ almost surely, $\boldsymbol{x}_{[2:p]}$ has mean zero and standard deviation one, and $||\boldsymbol{x}||_2\le X_{\max}\sqrt{p}$ \citep{ban2019big}.
The goal is to produce decisions that minimize predefined operational costs given the offline dataset $S_n$.

\subsubsection{Linear Regression.}
\label{subsec:CustLR}
If one uses an LR model to approximate the relationship between the decision and the features, i.e., $y_i = \sum_{j=1}^p{x_{i}^{j} \theta^j}$, where $\theta^j$'s are the weights of features, the learning objective is to determine the value of $\boldsymbol{\theta}$ and minimize the true risk:
\begin{equation}
\mathcal{R}^{\LReNVCa}(\boldsymbol{\theta})  =  \mathbb{E}\left[\inNVLoss(\boldsymbol{\theta}|(\boldsymbol{x},s))\right].
    \label{eq:LRObjTrue}
\end{equation}
However, since the decision-maker does not know the distribution of the true or censored variable, we use a sample-averaged proxy instead.
With the dataset $S_n$, the objective function for training an LR model with the $\varepsilon$-insensitive newsvendor cost, Equation~\eqref{eq:le1e2}, is given by
\begin{align}
    \hat{\mathcal{R}}^{\LReNVCa}(\boldsymbol{\theta};S_n) & = \frac{1}{n}\sum_{i=1}^n{\inNVLoss_i(\boldsymbol{\theta}|(\boldsymbol{x}_i,s_i))} \nonumber
    \\
    & = \frac{1}{n}\sum_{i=1}^n \left[ (1-\alpha)  (\sum_{j=1}^p{x_{i}^{j} \theta^j}-s_i-\varepsilon_1)^+ + \alpha (s_i+\varepsilon_2-\sum_{j=1}^p{x_{i}^{j} \theta^j})^+ \right]. \label{eq:LRObjInMae.w} 
\end{align}

To minimize Equation~\eqref{eq:LRObjInMae.w}, an LP problem can be formulated as Model~\ref{mod:LRLP}, where $u_i$ and $o_i$ are the underage and overage quantities, respectively.
\begin{align}
    \label{mod:LRLP}
    \tag{LP.$\varepsilon$LR}
    \min_{\boldsymbol{\theta}} \quad & {\frac{1}{n}\sum_{i=1}^n{\left[(1-\alpha) o_i + \alpha u_i \right]}}, \\
    \text{s.t.} \quad & u_i \geq \sum_{j=1}^p{x_{i}^{j} \theta^j} - s_i - \varepsilon_1, & \forall i \in \{1, \ldots, \nSamples\},\\
    & o_i \geq s_i + \varepsilon_2 - \sum_{j=1}^p{x_{i}^{j} \theta^j}, & \forall i \in \{1, \ldots, \nSamples\},\\
    & u_i \geq 0, & \forall i \in \{1, \ldots, \nSamples\},\\
    & o_i \geq 0, & \forall i \in \{1, \ldots, \nSamples\}.
\end{align}

Model~\ref{mod:LRLP} has $2n+p$ decision variables and $4n$ constraints and can be solved by commercial solvers such as CPLEX, though the scale of the problem increases with the sample size $n$. 
Therefore, we use the mini-batch \textit{gradient descent} (GD) method  (\citealp{robbins1951sgd, bottou2018optML,elmachtoub2022smart}) to train LR models with $\inNVLoss$ in the numerical experiments. 
Proposition~\ref{prop:LR-grad} provides the gradients with respect to the weights of features of the empirical loss \eqref{eq:LRObjInMae.w}.

\begin{proposition}[Gradients of $\inNVLoss$ in \LReNVC]
\label{prop:LR-grad}
The gradient with respect to the parameters $\theta^j$ in the LR model for the loss function $\inNVLoss_i(\boldsymbol{\theta}|(\boldsymbol{x}_i,s_i))$ is 
\begin{align}
    \frac{\partial \inNVLoss_i}{\partial \theta^j} = 
        \begin{cases}
            (1-\alpha)  x_{i}^{j}, & \sum_{j=1}^p{x_{i}^{j} \theta^j} > s_i+\varepsilon_1,\\
            -\alpha  x_{i}^{j}, & \sum_{j=1}^p{x_{i}^{j} \theta^j} < s_i+\varepsilon_2, \\
            0, & \text{otherwise}.
        \end{cases}
\end{align}
\end{proposition}

Denote by $S_{n_k}^{(k)}$ the batch dataset with batch size $n_k$ in the $k$-th iteration of the mini-batch GD method, and by $\eta$ the learning rate in each iteration. With the gradients given by Proposition~\ref{prop:LR-grad}, the update rules for $\theta^j$'s in the $(k+1)$-th iteration with the dataset $S_{n_{k+1}}^{(k+1)}$ are given by  
\begin{align}
\label{eq:LR-updateRule}
    \theta^{j (k+1)} = \theta^{j (k)} - \frac{\eta}{n_{k+1}}\sum_{i\in S_{n_{k+1}}^{(k+1)}}{x_{i}^{j}\left[\mathbb{I}(y_i^{(k)}> s_i+\varepsilon_1)(1-\alpha)+\mathbb{I}(y_i^{(k)}< s_i+\varepsilon_2)\alpha \right],}
\end{align}
where $\theta^{j (k)}$ is the value of $\theta^j$ in the $k$-th iteration, and $y_i^{(k)}$ is the output in the $k$-th iteration, i.e., $y_i^{(k)} = \sum_{j=1}^p{x_{i}^{j} \cdot \theta^{j (k)}}$.
Following the above update rules, we iteratively obtain the approximately optimal parameters $\boldsymbol{\theta}$ of the LR model.

In the high dimensional case where the number of features $p$ is large compared with the number of observations $n$, a regularization approach is used to automatically select features. Thus, we next introduce the \LReNVC ~with regularization, denoted by \LReNVCR. The objective function for \LReNVCR ~is given by
\begin{equation}
    \label{eq:LReNVCR}
    \min \hat{\mathcal{R}}^{\LReNVCa}(\boldsymbol{\theta};S_n) +\lambda ||\boldsymbol{\theta}||_k^2,
\end{equation}
where $\lambda>0$ is the regularization parameter and $||\boldsymbol{\theta}||_k^2$ denotes the $l_k$ norm of the vector $\boldsymbol{\theta}=[\theta^1,\ldots, \theta^p]^\intercal$. Corresponding to Model~\ref{mod:LRLP}, the problem in Equation \ref{eq:LReNVCR} can be formulated as
\begin{align}
    \label{mod:LReNVCR}
    \tag{LR-$\varepsilon$NVC-R}
    \min_{\boldsymbol{\theta}} \quad & {\frac{1}{n}\sum_{i=1}^n{\left[(1-\alpha) o_i + \alpha u_i \right]}}+\lambda ||\boldsymbol{\theta}||_k^2, \\
    \text{s.t.} \quad & u_i \geq \sum_{j=1}^p{x_{i}^{j} \theta^j} - s_i - \varepsilon_1, & \forall i \in \{1, \ldots, n\},\\
    & o_i \geq s_i + \varepsilon_2 - \sum_{j=1}^p{x_{i}^{j} \theta^j}, & \forall i \in \{1, \ldots, n\},\\
    & u_i \geq 0, & \forall i \in \{1, \ldots, n\},\\
    & o_i \geq 0, & \forall i \in \{1, \ldots, n\},
\end{align}

\subsubsection{Neural Networks.}
\label{subsec:CustNN}
\begin{figure}[!htbp]
    \centering
    \includegraphics[scale = 1.0]{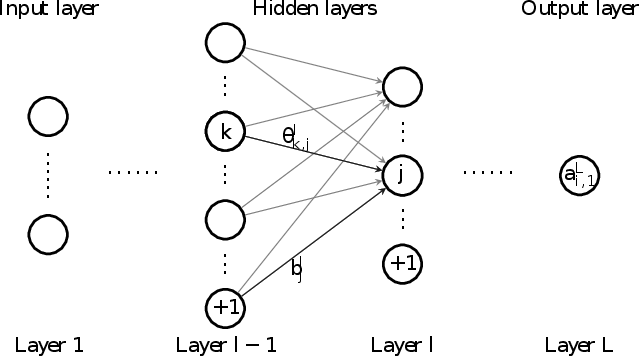}
    \caption{A Simple Neural Network}
    \label{Fig:SimpleNN}
\end{figure}

We depict a simple NN in Figure~\ref{Fig:SimpleNN}. For a detailed discussion of the NN, please refer to \cite{goodfellow2016dnn}. 
In Figure~\ref{Fig:SimpleNN}, the nodes between two adjacent layers are fully connected, and some layers are omitted for brevity. 
Suppose there are $\nLayers$ layers in the NN. We use the superscript $\layerN$ to denote the notation for layer $\layerN$. 
Denote by $\mathbb{N}^l$ the set of the index of nodes in layer $\layerN$.
Layer 1 is the input layer. A node in the input layer represents a feature.
Layers $\layerN$ ($\forall \ \layerN = 2, \ldots, \nLayers-1$) are hidden layers. 
The ``+1" nodes at the bottom of the \textit{hidden} layers represent bias units. 
Nodes other than the ``+1'' nodes in the hidden layer $\layerN$ take the linear combination ($\zeta_j^{\layerN}$) of the outputs of nodes in the previous layer as inputs, i.e., $\zeta_j^{\layerN} = \sum_{k\in \mathbb{N}^{l-1}}{\theta_{j,k}^\layerN a_k^{\layerN-1}} +b_j^{\layerN}$, and output $a_j^\layerN = \sigma^\layerN(\zeta^\layerN_j)$ to layer $\layerN+1$, where $\sigma^\layerN(\cdot)$ is the sigmoid activation function in layer $\layerN$, $\theta_{j,k}^\layerN$ represents the weight between node $k$ in layer $\layerN-1$ and node $j$ in layer $\layerN$, and $b_j^{\layerN}$ represents the bias for node $j$ in layer $\layerN$.
Layer $\nLayers$ is the output layer. We use one node in the output layer.
This node's output and the NN's output is $a_1^\nLayers$.

An NN is trained to determine the weight vector $\boldsymbol{\theta}$ and the bias vector $\boldsymbol{b}$ of the network so that the outputs from the NN have almost the lowest value of the loss function. 
We use the back-propagation algorithm \citep{rumelhart1986backpropagating} and SGD to train the NNs.

Before we derive the gradients of the $\varepsilon$-insensitive loss function with respect to the parameters (i.e., $\boldsymbol{\theta}$ and $\boldsymbol{b}$), we set some notations.
For the $i$-th observation, let 
\begin{equation}
    \delta_j^{i,\layerN} = \frac{\partial \inNVLoss_i}{\partial \zeta_j^{i,\layerN}} =\frac{\partial \inNVLoss_i}{\partial a_j^{i,\layerN}}\frac{\partial a_j^{i,\layerN}}{\partial \zeta_j^{i,\layerN}} 
    = \frac{\partial \inNVLoss_i}{\partial a_j^{i,\layerN}} \sigma'(\zeta_j^{i,\layerN}).
\end{equation}
Then, for the output layer $\nLayers$ with activation function $\sigma^\nLayers(\tau) = \tau$, we have
\begin{equation}
    \delta_1^{i,\nLayers} = \frac{\partial \inNVLoss_i}{\partial \zeta_1^{i,\nLayers}} =\frac{\partial \inNVLoss_i}{\partial a_1 ^{i,\nLayers}}
        =
        \begin{cases}  
            \alpha, &y_i\leq s_i+\varepsilon_2, \\
            1-\alpha, &y_i\geq s_i+\varepsilon_1, \\
            0, &y_i\in(s_i+\varepsilon_2,s_i+\varepsilon_1).
        \end{cases}
\end{equation}
By mathematical induction, we have
\begin{align}
    \delta_k^{i,\layerN-1} &= \frac{\partial \inNVLoss_i}{\partial \zeta_k^{i, l-1}} =
    \sum_{j \in \mathbb{N}^l}{\frac{\partial \inNVLoss_i}{\partial a_j^{i,\layerN}}}\frac{\partial a_j^{i,\layerN}}{\partial \zeta_k^{i,\layerN-1}} 
    = \left( \sum_{j \in \mathbb{N}^l} {\theta_{j,k}^{\layerN}\delta_j^{i,\layerN}} \right) \sigma'(\zeta_k^{i,\layerN-1}), & \quad \forall k \in \mathbb{N}^{l-1}, \forall l = 2, \ldots, \nLayers.
\end{align}

Using the notation of $\delta_j^{i,l}$, Proposition \ref{prop:DNNRule} provides the gradients of the $\varepsilon$-insensitive newsvendor cost, i.e., Equation~\eqref{eq:le1e2}, for the weights of the network. With the SGD algorithm, our NN algorithm uses the gradients provided in Proposition~\ref{prop:DNNRule} to update the weights and bias of the network iteratively.
\begin{proposition}[Gradients of $\inNVLoss$ in \NNeNVC]
\label{prop:DNNRule}
The gradient for the weights of the neural network for the $\varepsilon$-insensitive newsvendor cost \eqref{eq:le1e2} is
\begin{equation}
    \frac{\partial \inNVLoss_i}{\partial \theta_{j,k}^\layerN} = a_k^{i,\layerN-1} \delta_j^{i,\layerN},
\end{equation}
and the gradient for the bias of the network is
\begin{equation}
    \frac{\partial \inNVLoss_i}{\partial b_{j}^\layerN} = \delta_j^{i,\layerN}.
\end{equation}
\end{proposition}

\subsection{Properties of the Custom Learning Algorithms}
\label{sec:prop}

We denote the LR with the $\varepsilon$-insensitive newsvendor cost by \LReNVC, \LReNVCR ~and the NN \NNeNVC.
This section proves that the proposed \LReNVC, \LReNVCR ~and \NNeNVC ~are uniformly stable and provides the generalization error bounds. 

The generalization performance of an ML algorithm is one of the key issues in ML. It can explain why the ML algorithm can learn from the observed limited sample and generalize to the unobserved sample. 
A prominent theory of learnability is the theory of uniform convergence of empirical quantities to their mean (see \cite{vapnik1982estimation} for example). 
\cite{bousquet2002stability} introduce the concept of uniform stability and show that it is a sufficient condition for learnability. 
\cite{shalev2010learnability} show that stability is the key necessary and sufficient condition for learnability, rather than uniform convergence in the general learning setting. This section will first demonstrate the uniform stability and generalization of \LReNVC ~and \LReNVCR ~using the theories proposed by \cite{bousquet2002stability}.

Since the concept of uniform stability developed by \cite{bousquet2002stability} is valid for deterministic learning algorithms, it is not suitable to derive upper bounds for the generalization error of the NNs trained by SGD. 
We then introduce the concept of uniform stability to the randomized learning algorithm proposed by \cite{hardt2016train} and develop the generalization guarantee for \NNeNVC ~trained by multi-pass SGD. The multi-pass SGD algorithm runs SGD for multiple passes (epochs) over the dataset.

To obtain the stability results of our proposed algorithm, we apply Hoeffding's inequality \citep{Hoeffding1963prob}, and we need a bound on the value of a random variable. Thus, we assume that there is a bound on the unknown true value, i.e., $d\in[0,\bar{D}]$, and thus a bound on the corresponding censored observation, $s \in [0,\bar{D}]$. And we show that the cost function in our proposed algorithms has a tight uniform bound in Lemma \ref{lem:CostBound}.

\begin{lemma}[Tight uniform bound on $\inNVLoss$]
\label{lem:CostBound}
The custom $\varepsilon$-insensitive operational cost function 
    \begin{align}
    \inNVLoss(s,y(\boldsymbol{x})) &= (1-\alpha)(y-s-\varepsilon_1)^++\alpha(s+\varepsilon_2-y)^+ , & \varepsilon_1> \varepsilon_2\geq 0,
    \end{align}
is bounded by $(\alpha \vee (1-\alpha))(\bar{D}+\varepsilon_2)$, which is tight in the sense that:
\begin{equation}
\sup_{(\boldsymbol{x},s)\in \mathcal{X} \times \mathcal{D}}{|\inNVLoss|} = (\alpha \vee (1-\alpha))(\bar{D}+\varepsilon_2).
\end{equation}
\end{lemma}

In this paper, $\alpha \vee (1-\alpha) \triangleq \max\{\alpha, 1-\alpha\}$, and $\alpha \wedge (1-\alpha) \triangleq \min\{\alpha, 1-\alpha\}$. Lemma \ref{lem:CostBound} provides a tight uniform bound for the $\varepsilon$-insensitive operational cost function $\inNVLoss$. The upper bound for the $\varepsilon$-insensitive cost function increases with the upper bound of the true value $\bar{D}$. 
It is intuitive that only one of the insensitive parameters, $\varepsilon_2$, is included in the upper bound of $\inNVLoss$.

The Lipschitz property of a loss function is the foundation for generalization error analysis. Lemma \ref{lem:lipschitz} provides that the $\varepsilon$-insensitive operational cost function $\inNVLoss$ is $(\alpha \vee(1-\alpha))$-Lipschitz.

\begin{lemma}
\label{lem:lipschitz}
The $\varepsilon$-insensitive operational cost function $\inNVLoss(s,\cdot)$ satisfies the Lipschitz condition with the Lipschitz constant $(\alpha \vee(1-\alpha))$ for every $s$.
\end{lemma}

Based on the definition of uniform stability in \citet[Definition 6]{bousquet2002stability}, we have the stable property for \LReNVC ~in Proposition~\ref{prop:LR-stable}.

\begin{proposition}[Uniform stability of \LReNVC]
\label{prop:LR-stable}
The learning algorithm \LReNVC ~is uniformly stable with the stability parameter  
\begin{align}
\xi_n = \frac{p}{n} \frac{(\alpha \vee (1-\alpha))^2}{(\alpha \wedge (1-\alpha))} (\bar{D}+\varepsilon_2).
\end{align}
\end{proposition}

Proposition~\ref{prop:LR-stable} gives a bound on the stability of \LReNVC ~to a random dataset. Proposition \ref{prop:LR-stable} also shows that \LReNVC ~has a very strong stability property such that deleting an observation in the training dataset does not change much of its decision cost on the training set. The algorithm's stability means the stability of its decision, which implies that the expected cost of the decision is not sensitive to the change in the dataset. It is a desirable property indicating how well the decision performs on new observations. 
From Proposition~\ref{prop:LR-stable}, for a fixed weighted parameter $\alpha$ and a fixed number of features $p$, the stability parameter of \LReNVC  ~is scaled as $\frac{1}{n}$.

After obtaining the uniform stability property, we provide the generalization bound of \LReNVC ~in Theorem \ref{theorem:LR-GenBound}. 
The generalization bound of an algorithm captures the variance of the in-sample decision, and the bound of generalization error leads directly to provable guarantees on the excess population risk of a learning algorithm~\citep{bassily2020stability}.

\begin{theorem}[Generalization bound for \LReNVC]
\label{theorem:LR-GenBound}
Let $\hat{\boldsymbol{\theta}}$ denote the parameters produced by \LReNVC ~with the sample set $S_n$. With probability at least $1-\delta$ over the random draw of the sample set $S_n$, we have the following tight generalization bound:
\begin{align}
\frac{\mathcal{R}^{\LReNVCa}(\hat{\boldsymbol{\theta}})-\hat{\mathcal{R}}^{\LReNVCa}(\hat{\boldsymbol{\theta}};S_n)}{(\alpha \vee (1-\alpha)) (\bar{D}+\varepsilon_2)} \leq \frac{2p}{n} \frac{\alpha \vee (1-\alpha)}{\alpha \wedge (1-\alpha)} + \left(4p \frac{\alpha \vee (1-\alpha)}{\alpha \wedge (1-\alpha)}+1\right) \sqrt{\frac{\ln{1/\delta}}{2n}}.
\end{align}
\end{theorem}

Theorem \ref{theorem:LR-GenBound} gives a tight upper bound on the generalization error of \LReNVC, which bounds the difference between the training error and the true error for the in-sample decision. 
For a fixed weighted parameter $\alpha$, the generalization error scales as $O(p/\sqrt{n})$, which is the best finite-sample bounds for this problem as described in \cite{ban2019big}. 
Note also that, in many circumstances, the sample size is greater than the feature number, and $p/\sqrt{n}$ decreases as sample size $n$ increases. In such cases, overfitting is not an issue for \LReNVC. 
In addition, the generalization bound also contains an upper bound of the true value, which ensures that the generalization bound is not scale-invariant. 
It is worth noting that the upper bound of the generalization error is related to the insensitive parameter. The larger the insensitive parameter, the larger the bound. This is because the larger the insensitive parameter is, the greater the tolerance for the predicted value when training the model.

Analogously, we derive the stability and learnability results for the custom LR algorithm with regularization \LReNVCR.
\begin{proposition}[Uniform stability of \LReNVCR]
\label{prop:LR-R-stable}
The learning algorithm \LReNVCR ~is uniformly stable with the stability parameter  
\begin{align}
\xi_n^r = \frac{(\alpha \vee (1-\alpha))^2 X_{\max}^2 p}{2 n \lambda}
\end{align}
\end{proposition}

\begin{theorem}[Generalization bound for \LReNVCR]
\label{theorem:LR-R-GenBound}
Let $\hat{\boldsymbol{\theta}}^r$ denote the parameters produced by \LReNVCR ~with the sample set $S_n$. With probability at least $1-\delta$ over the random draw of the sample set $S_n$, we have the following tight generalization bound:
\begin{equation}
    \begin{aligned}
    \mathcal{R}^{LR-\varepsilon NVC-R}(\hat{\boldsymbol{\theta}}^r)-
    \hat{\mathcal{R}}^{LR-\varepsilon NVC-R}(\hat{\boldsymbol{\theta}}^r;S_n)
    &\leq \frac{\alpha \vee (1-\alpha)^2}{X_{\max}^{-2}}\frac{p}{n \lambda} + \\
    &\left(\frac{2 (\alpha \vee (1-\alpha))^2}{X_{\max}^{-2}}\frac{p}{\lambda}\right)\sqrt{\frac{\ln{2/\delta}}{2n}}+\\
    &\left((\bar{D}+\varepsilon_2)(\alpha \vee (1-\alpha))\right) \sqrt{\frac{\ln{2/\delta}}{2n}}.
    \end{aligned}
\end{equation}
\end{theorem}

Next, we propose the uniform stability and generalization bound for \NNeNVC ~trained by $K$-pass SGD.

Before we derive the generalization bound for \NNeNVC, we describe two foundational concepts. Recall that the dataset is denoted by $S_n \in \mathcal{X} \times \mathcal{D}$, and the decision variable is denoted by $y \in \mathcal{Y} \subseteq \mathbb{R}$. We use $S_n \simeq S'_n$ to denote that the two datasets $S_n$ and $S'_n$ differ only in a single point.
\begin{enumerate}
    \item 
    A \textit{stochastic optimization algorithm} is a randomized mapping $\mathcal{A}: \mathcal{X} \times \mathcal{D} \rightarrow \mathcal{Y}$. 
    \item
    Given an algorithm $\mathcal{A}$ and datasets $S_n\simeq S'_n$, the \textit{uniform argument stability} random variables \citep{bassily2020stability} are defined as 
  \begin{align}
  \delta_{\mathcal{A}}(S_n,S_n'):=|\!|\mathcal{A}(S_n)-\mathcal{A}(S_n')|\!|.
  \end{align}
  The randomness of the UAS here is due to any possible internal randomness of $\mathcal{A}$. For any $\gamma$-Lipschitz function $\mathcal{L}$, we have that $\mathcal{L}(s,\mathcal{A}(S_n))-\mathcal{L}(s,\mathcal{A}(S_n'))\leq \gamma \delta_{\mathcal{A}}(S_n,S_n')$. Hence, upper bounds on UAS can be easily transformed into upper bounds on uniform stability.
\end{enumerate}

Following \cite{hardt2016train,bassily2020stability}, and \cite{akbari2021how}, we obtain the UAS bound and the generalization bound for \NNeNVC ~trained by $K$-pass SGD.

\begin{proposition}[Bound on UAS for \NNeNVC ~(SGD)]
\label{prop:DNN-UASBound}
Suppose the learning rate used in SGD is a constant $\eta$. The uniform argument stability of \NNeNVC ~trained by $K$-pass SGD with a fixed permutation is bounded with probability 1 as 
\begin{align}
    \sup_{S_n\simeq S_n'}\delta(S_n,S_n')\leq 2 (\alpha \vee (1-\alpha))(\eta\sqrt{nK}+2\eta K).
\end{align}
\end{proposition}

Proposition \ref{prop:DNN-UASBound} gives the bound on the uniform argument stability random variable of \NNeNVC ~trained by $K$-pass SGD.
For a fixed weighted parameter $\alpha$, the bound is scaled as $O(\eta \sqrt{nK})$ if $n>K$, and $O(\eta K)$ otherwise. 
In both cases, the algorithm's stability can be enhanced by setting reasonable learning rates $\eta$, for example, setting the learning rate as $\frac{1}{n}$ if $n>K$ and $\frac{1}{K^2}$ otherwise. 
In other words, if the sample size is small, we can choose a smaller learning rate and execute the SGD for more passes for a more stable result.

\begin{theorem}[Generalization bound for \NNeNVC ~(SGD)]
\label{theorem:dnnGBound}
Let $f_S$ denote the model produced by learning algorithm \NNeNVC ~with the sample set $S_n$. Suppose SGD is run for $K$ passes (i.e., $n K$ iterations) with a fixed permutation and a constant learning rate $\eta$ to find $f_S$. Then, there exists $c>0$, such that with probability $1-\rho$, we have the following generalization bound: 
\begin{align}
\label{eq:ubDNNGB}
\begin{split}
& |\mathcal{R}^{\NNeNVCa}(f_S)-\hat{\mathcal{R}}^{\NNeNVCa}(f_S;S_n)|\\
&\leq c (\alpha \vee (1-\alpha))^2 (\eta\sqrt{n K}+2K\eta)\log(n)\log(n/\rho)+ c(\alpha \vee (1-\alpha)) \bar{D}\sqrt{log(1/\rho)/n}.
\end{split}
\end{align}
\end{theorem}

Theorem \ref{theorem:dnnGBound} gives an upper bound on the generalization error of \NNeNVC ~trained by $K$-pass SGD. For a fixed weighted parameter $\alpha$, setting the learning rate $\eta = \frac{\bar{D}}{(\alpha \vee (1-\alpha))n\sqrt{K}}$, the generalization error of \NNeNVC ~is bounded by $O\left(\frac{1}{\sqrt{n}}\log(n)\log(\frac{n}{\rho})\right)$ with probability $1-\rho$. Also, the generalization bound of \NNeNVC ~trained by $K$-pass SGD is scaled with the upper bound of the true value $\bar{D}$.

\section{Numerical Experiments}
\label{sec:numExp}

In this section, we test the effectiveness of our proposed decision-making approaches using a newsvendor problem. The newsvendor model is a fundamental stochastic inventory model dating back to \cite{edgeworth1888mathematical}; see \cite{zipkin2000foundations} for textbook discussions. In this section, we consider an extension of the newsvendor model, a repeated multi-feature newsvendor model for multiple products \citep{oroojlooyjadid2020applying}.
We create synthesized datasets based on real-world data. 
The synthesized datasets provide true underlying distributions that are typically unavailable to managers and allow us to measure the loss our proposed methods could recover by considering censoring.
In the following, we introduce the problem setup, the data generation process, the experimental design, and the results.

\subsection{Problem Description}

A retailer sells $M$ perishable products, indexed by $m \in \{1, \ldots, M\}$.  
The demands for these products are assumed to be independent. 
The retailer makes the order decision at the beginning of a period and sells them during the period. 
We assume there is no lead time. 
At the end of the period, unsold goods must be discarded, and the retailer incurs an overage cost. 
If the goods run out before the end of the period, the retailer incurs an underage cost.

The retailer needs to determine the optimal order quantities given the offline dataset of $T$ periods and minimize the operational cost.
Although most extant studies assume demand to be fully observable and the realized sales to be equal to demand, there are myriad business contexts in which sales are only the uncensored part of the full demand.
In this study, we propose solution methods using $\varepsilon$-insensitive operational costs, as shown in Equation~\eqref{eq:in-nvc} for the newsvendor with censored demand data:
\begin{align}
    & \frac{1}{TM}\sum_{t=1}^{T}\sum_{m=1}^M \left[ c_u (s_{m,t} + \varepsilon_2 - q_{m,t}(\boldsymbol{x}_{m,t}))^+ + c_o (q_{m,t}(\boldsymbol{x}_{m,t}) - s_{m,t} - \varepsilon_1)^+ \right], & \varepsilon_1>\varepsilon_2 \geq 0, \label{eq:in-nvc}
\end{align}
where
\begin{itemize}
    \item $c_u$ is the unit underage cost;
    \item $c_o$ is the unit overage cost;
    \item $s_{m,t}$ is product $m$'s sales in period $t$; 
    \item $\boldsymbol{x}_{m,t}$ is the feature vector of product $m$ in period $t$; and
    \item $q_{m,t}$ is the retailer's order decision of product $m$ in period $t$ and is a function of $\boldsymbol{x}_{m,t}$.
\end{itemize}

Defining the operational loss as Equation~\eqref{eq:in-nvc} indicates that when $q_{m,t}$ is set lower than sales $s_{m,t}$, since the sales could be the censored demand, the actual loss might be more than $c_u (s_{m,t} - q_{m,t}(\boldsymbol{x}_{m,t}))$. We denote the amount by $c_u \varepsilon_2$ and estimate its value using ERM.
On the other hand, when $q_{m,t}$ is set higher than sales $s_{m,t}$, the overage cost could be smaller than $c_o (q_{m,t}(\boldsymbol{x}_{m,t}) - s_{m,t})$. We denote the amount by $c_o \varepsilon_1$ and estimate it using ERM.
Lastly, we normalize the loss values by defining $\alpha = \frac{c_u}{c_u+c_o}$ and transform Equation~\eqref{eq:in-nvc} into the following form: 
\begin{align}
    \mathcal{L}^{\varepsilon NVC} \! &= \! \frac{1}{TM}\sum_{t=1}^{T}\sum_{m=1}^M \left[ \alpha (s_{m,t} +  \varepsilon_2 -  q_{m,t}(\boldsymbol{x}_{m,t}))^+ \!+ \! (1 \!- \! \alpha) (q_{m,t}(\boldsymbol{x}_{m,t}) \! - \! s_{m,t} -  \varepsilon_1)^+ \right], \! \!  &\varepsilon_1 \! > \!\varepsilon_2 \! \geq \! 0. \label{eq:in-nvc-norm}
\end{align}

In our numerical study, we utilize Equation~\eqref{eq:in-nvc-norm} to train the learning algorithms and obtain the newsvendor order quantities.

\subsection{Data}
The decision maker makes order decisions based on the offline dataset $S_n = \{\boldsymbol{x}_i, s_i \}_{i = 1}^n$, and in order to study the censoring phenomena and evaluate the effectiveness of our proposed algorithms, we should use demand data to calculate the out-of-sample newsvendor cost in the test set. Thus, the generated simulation dataset consists of the feature data $\boldsymbol{x}$, the sales data $s_i$, and the demand data $d_i$.

We prepare our synthesized data based on a real-world dataset of Corporaci\'{o}n Favorita's transactions from Kaggle (2017).
We assume that the demand generation process follows a linear model with a norm noise, i.e., $D=\boldsymbol{\beta}^\intercal \boldsymbol{x}+D^r$, where $\boldsymbol{\beta}$ are the coefficients of features $\boldsymbol{x}$, and $D^r\sim \mathcal{N}(\mu,\sigma^2)$.
We choose the feature available in the real-world dataset, and estimate the parameter $(\boldsymbol{\beta},\mu,\sigma)$ with the selected store's transaction records. The available features are \textit{Month of Year}, \textit{Day of Week}, and \textit{Product Category}.
We implement dummy encoding for categorical feature variables, after which the dimension of the feature variables $\boldsymbol{x}$ is 26, including the intercept.
The estimated result of the LR model is summarized in Table \ref{tab:LRstore10} of Appendix \ref{app:storeSelect}.
We estimate the mean $\mu$ and standard deviation $\sigma$ of the random noise $D^r$ using the residuals of the LR model for the selected store, obtaining a mean of 0 and a standard deviation of 46.57. The details are listed in Appendix \ref{app:storeSelect}.
Note that, our proposed framework does not require the true demand to conform to such an LR model.
We know this LR model is trained with observed sales and not the demand of the perishable categories of interest. Nevertheless, this allows us to study the censoring phenomena and measure the effectiveness of our proposed methods. In related literature, \cite{oroojlooyjadid2020applying} synthesize demand from a single probability distribution with no noise, while \cite{beutel2012safety} generate demand by assuming it has a linear relationship with an exogenous variable, price, plus an error with the standard normal distribution, and \cite{ban2019big} assume a linear relationship between demand and its features.

We use the calendar days from January 1, 2016, to June 30, 2017. There are 9 perishable product categories in the real-world dataset. For each product category and on each day, we have readily the value of each feature, namely, \textit{Month of Year}, \textit{Day of Week}, and \textit{Product Category}. Plugging in the feature values in Table \ref{tab:LRstore10} of Appendix \ref{app:storeSelect}, we have the deterministic part of the demand, denoted as $D^c=\hat{\boldsymbol{\beta}}^\intercal \boldsymbol{x}$. Then, we add a random noise, $D^r$, from a normal distribution that we estimate from the residuals of the LR. The normal distribution is $\mathcal{N}(0, 46.57^2)$. The synthesized demand is $d_i = d^c_i + d^r_i$. We generate the synthesized demand data independently, which is consistent with the assumptions in the problem description that there is no autocorrelation in time or dependency across product categories' demand. We use 10 random seeds to generate 10 sets of synthesized demand. We run each experiment 10 times on a different dataset and obtain 10 sets of results to ensure that our results are robust.

Next, we synthesize sales assuming that the newsvendor's historical order quantities are the mean of the demand, i.e., $Q=D^c$. \cite{schweitzer2000decision} and \cite{ren2013overconfidence} point out that in practice, there is a pull-to-center effect in the newsvendor's ordering decision, and the order quantity tends to approach the average demand. Moreover, setting the ordering strategy as the demand mean can exploit various censoring degrees of sales data under different cost parameters. We set the sales volume as the smaller one between the order quantity and the synthesized demand.

Since there is a sequence of data acquisition in the actual situation, we take the transactional data in 2016 as a training set with 3286 observations and that in the first half of 2017 as a test set with 1629 observations. During model training, we use the feature data and the sales data but not the demand data in the training set. We only use the demand data in the test set to measure the performance of our algorithms in dealing with censoring.

\subsection{Experiment Design}

Next, we utilize the synthesized datasets and train LR and NNs to learn the newsvendor's order decisions.
We compare the solutions of our methods (\LReNVCR ~and \NNeNVC) to two existing approaches: EAS \citep{oroojlooyjadid2020applying} (\LRMSE ~and \NNMSE) and IEO \citep{ban2019big} (\LRNVC ~and \NNNVC).
\LReNVCR ~and \NNeNVC ~use the custom newsvendor cost in Equation~\eqref{eq:in-nvc-norm} for training LR and NNs. 
The EAS approaches take the demand predictions as the order quantities and use the following MSE cost as the training loss:
\begin{equation}
    \mathcal{L}^{MSE} = \frac{1}{TM} \sum_{t=1}^T \sum_{m=1}^{M} (y_{m,t} - s_{m,t})^2. \label{eq:MSELoss}
\end{equation}
The IEO approaches utilize the predictive model to optimize the operational cost, i.e., the newsvendor cost, and solve for the optimal order quantities. Thus, the IEO approaches use the following standard newsvendor cost to train LR and NNs:
\begin{align}
    \mathcal{L}^{NVC} &= \frac{1}{TM} \sum_{t=1}^T \sum_{m=1}^{M} \left[ \alpha (s_{m,t} - y_{m,t})^+ + (1-\alpha) (y_{m,t} - s_{m,t})^+ \right], & \alpha \in (0,1). \label{eq:NVLoss}
\end{align}
However, neither the EAS nor the IEO approaches account for censored observations.
Without the point-of-sales data and the inventory data in the offline dataset, these are the state-of-the-art methods we know of for this problem.

During training, the labels are the observed sales, since the true demand is assumed to be unobservable. 
Meanwhile, for testing, the true demand is used for computing the out-of-sample newsvendor cost, which is defined as
\begin{align}
    NVCost &= \frac{1}{TM} \sum_{t=1}^T \sum_{m=1}^{M} \left[ \alpha (d_{m,t} - y_{m,t})^+ + (1-\alpha) (y_{m,t} - d_{m,t})^+ \right], & \alpha \in (0,1).\label{eq:nvcost}
\end{align}

To evaluate the models' performance, we also compare the order quantities obtained by the models to the optimum, denoted by $Q^*$. $Q^*$ is obtained by optimally solving the newsvendor model with true demand distribution. Specifically, denoting $\Phi(\cdot)$ as the distribution function of $D^r$, the optimal order quantity under our experiment settings is defined as
\begin{equation}
    Q^*_{|\alpha} = \hat{\beta} \boldsymbol{X} + \Phi^{-1}(\alpha). \label{eq:qopt}
\end{equation}
We compute and compare RMSE based on the optimal order quantities ${Q_{m,t}^*}$ on the test set, denoted as $RMSE^Q$ and defined as
\begin{equation}
    RMSE^{Q} =  \sqrt{\frac{1}{TM} \sum_{t=1}^T \sum_{m=1}^{M} ({Q_{m,t}^*} - y_{m,t})^2 }. \label{eq:rmseq}
\end{equation}
Note that different $\alpha$'s lead to different ${Q_{m,t}^*}$'s and different $RMSE^Q$'s. For brevity, $\alpha$ is omitted from Equation~\eqref{eq:rmseq}, though we calculated $RMSE^Q$'s for different $\alpha$'s.

Please refer to Appendix \ref{app:training} for details of our training processes.
All ML models were built in \texttt{Python 3.8}, and all NN-based models were implemented using a popular neural network library, \texttt{Keras 2.6}.
All experiments were run in Windows 11 (21H2) on an Intel(R) Core(TM) i7-9700 CPU @ 3.00GHz processor with 8GB RAM.

\subsection{Results}

This section discusses the experimental results of the six algorithms, namely, \LRMSE, \LRNVC, \LReNVCR, \NNMSE, \NNNVC, and \NNeNVC. We report the average results over 10 datasets.

\subsubsection{Fitting Times.}

We first compare and report the fitting time of each algorithm.
The fitting time is the time for estimating the parameters and is not the time for tuning hyperparameters.
Table~\ref{tab:trainingTime} reports the average fitting time of the six algorithms on the 10 random datasets.

Among the three LR-based algorithms, for each $\alpha$, \LRMSE ~takes the least  time, while the fitting times of \LRNVC ~and \LReNVCR ~are larger, but within a reasonable range.  This is because we use the \texttt{scikit-learn 1.0.1} library to train \LRMSE, and we code the training process by ourselves for \LRNVC ~and \LReNVCR ~to incorporate the custom loss function. 
In the meantime, the fitting times of the three NN-based algorithms are around the same.

The fitting times of \LReNVCR ~(\NNeNVC) are about the same as those of \LRNVC ~(\NNNVC).
Therefore, our proposed algorithms can be implemented in practice, and the custom cost function does not lead to extra time.

\begin{table}[!htbp]
\centering
\footnotesize
\setlength{\tabcolsep}{4mm} 
\renewcommand\arraystretch{1}  
\caption{Average Fitting Time (Seconds)}
\label{tab:trainingTime}
\begin{tabular}{@{}ccccccc@{}}
\toprule
 & \LReNVCR & \LRNVC & \LRMSE & \NNeNVC & \NNNVC & \NNMSE \\ \hline
$\alpha=0.55$ & 7.92  & 7.48  & 0.015  & 10.68 & 12.20 & 12.49 \\
$\alpha=0.65$ & 8.18  & 7.69  & 0.0037 & 9.72  & 12.92 & 12.20 \\
$\alpha=0.75$ & 7.58  & 9.65  & 0.0043 & 10.19 & 12.70 & 12.00 \\
$\alpha=0.85$ & 6.41  & 10.79 & 0.0038 & 11.20 & 12.80 & 12.04 \\
$\alpha=0.95$ & 10.05 & 11.01 & 0.0037 & 12.86 & 12.69 & 12.12 \\ \bottomrule
\end{tabular}
\end{table}

\subsubsection{Out-of-Sample Newsvendor Costs.}

\begin{figure}[!htbp]
\centering
\subfigure[LR]{
\includegraphics[scale = 0.5]{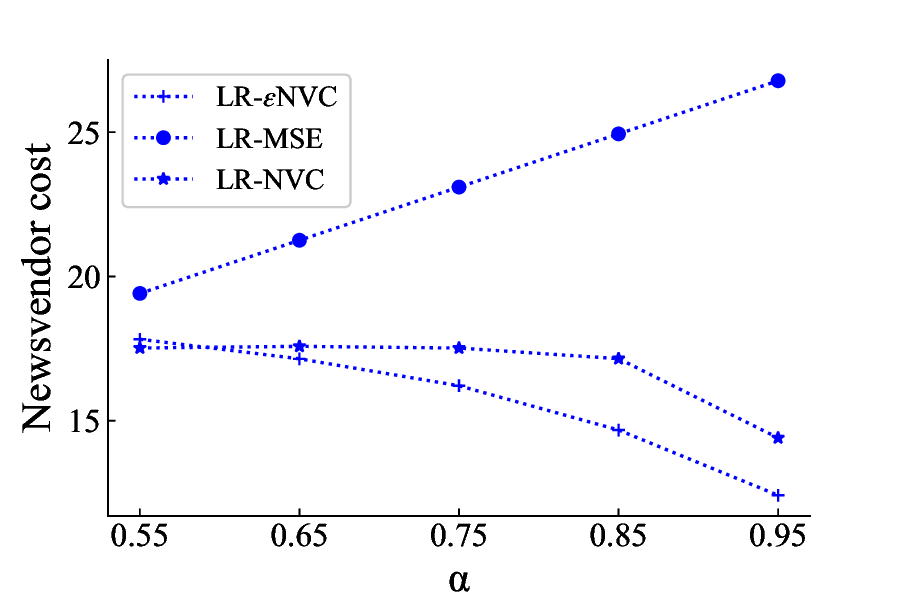}}
\subfigure[NN]{
\includegraphics[scale=0.5]{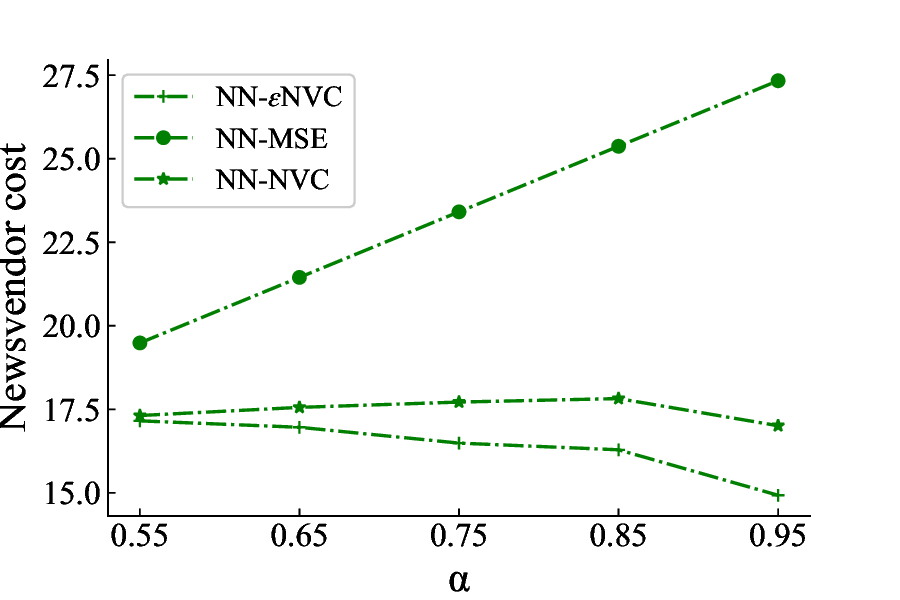}}
\caption{Out-of-Sample Average Newsvendor Costs}
\label{fig:testNVcost}
\end{figure}

To evaluate the decisions produced by the six algorithms, we compare their out-of-sample average newsvendor costs. 
We calculate the average newsvendor cost in Equation~\eqref{eq:NVLoss} on test sets using the actual demand as $d_{m,t}$ and the order quantities predicted by the models as $y_{m,t}$. 
Such newsvendor costs could reflect the actual empirical costs when implementing certain decision-making algorithms.
The out-of-sample average newsvendor costs of LR-based and NN-based models are shown in Figure \ref{fig:testNVcost}. 

First, for $\alpha>0.55$, the out-of-sample newsvendor costs of \LRNVC ~and \NNNVC ~are smaller than those of \LRMSE~and \NNMSE.
The newsvendor's goal is to obtain the optimal order quantity, not to predict demand. 
Therefore, the algorithms that minimize the newsvendor cost and the decision error, i.e., \LRNVC ~and \NNNVC, outperform the ones that minimize the demand prediction error, i.e., \LRMSE ~and \NNMSE.
This result is consistent with the findings in \cite{elmachtoub2022smart}. 

Second, the out-of-sample average newsvendor costs of \LReNVCR ~and \NNeNVC ~are less than those of \LRNVC ~and \NNNVC.
The data we have for tuning the algorithms are the sales instead of the demand. 
In such cases, using the standard newsvendor cost would take the sales as demand, ignoring the censoring.
By attaching no cost to order quantities in the range of $(s_i + \varepsilon_2, s_i + \varepsilon_1)$, the $\varepsilon$-insensitive newsvendor costs encourage the algorithms to produce order quantities that are larger than the observed sales.

\begin{figure}[!htbp]
\centering
\includegraphics[scale = 0.6]{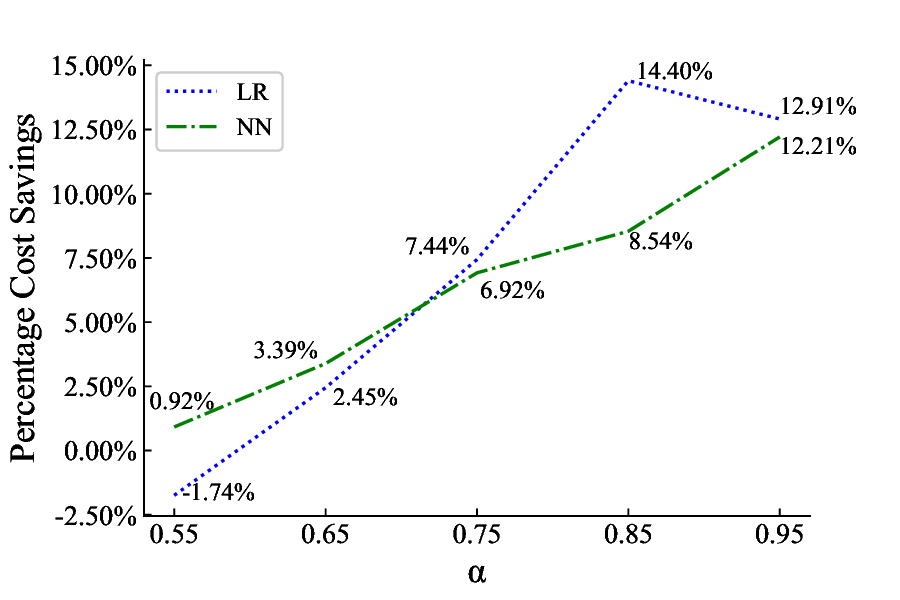}
\caption{Mean Percentage Cost Savings of \LReNVCR ~(\NNeNVC) ~over \LRNVC ~(\NNNVC)}
\label{fig:percentage}
\end{figure}
We present the average cost-saving percentages of \LReNVCR ~(\NNeNVC) over \LRNVC ~(\NNNVC) for each $\alpha$, shown in Figure \ref{fig:percentage}.
In general, the average cost-saving percentage increases with $\alpha$.
This indicates that the algorithms with the $\varepsilon$-insensitive operational cost are more effective when the censoring is severe. 
The cost-saving percentage of \LReNVCR ~over \LRNVC ~can be as high as 14.4\% ($\alpha = 0.85$), and that of \NNeNVC ~over \NNNVC, 12.21\% ($\alpha = 0.95$).

\subsubsection{Order Quantities.}

\begin{figure}[!htbp]
\centering
\subfigure[LR]{
\includegraphics[scale = 0.5]{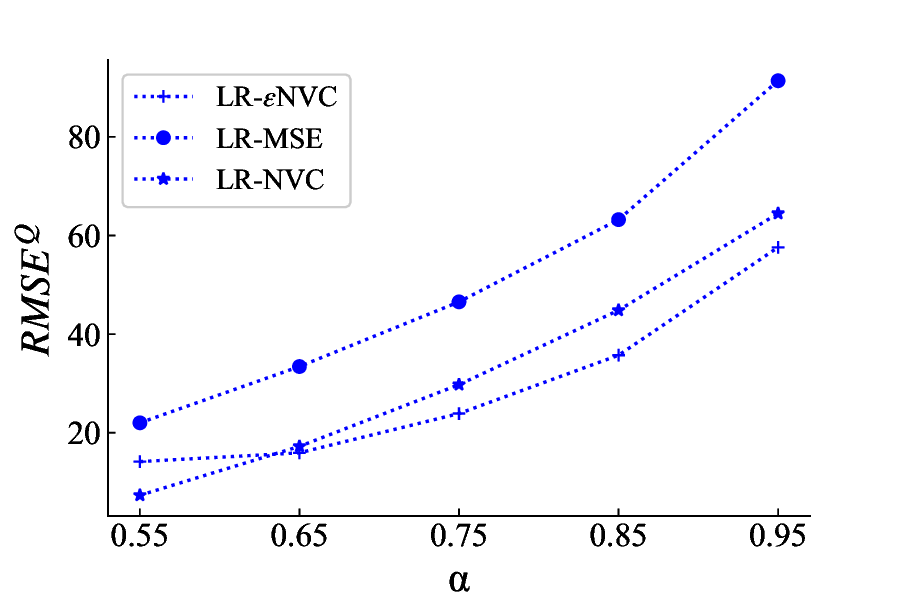}}
\subfigure[DNN]{
\includegraphics[scale=0.5]{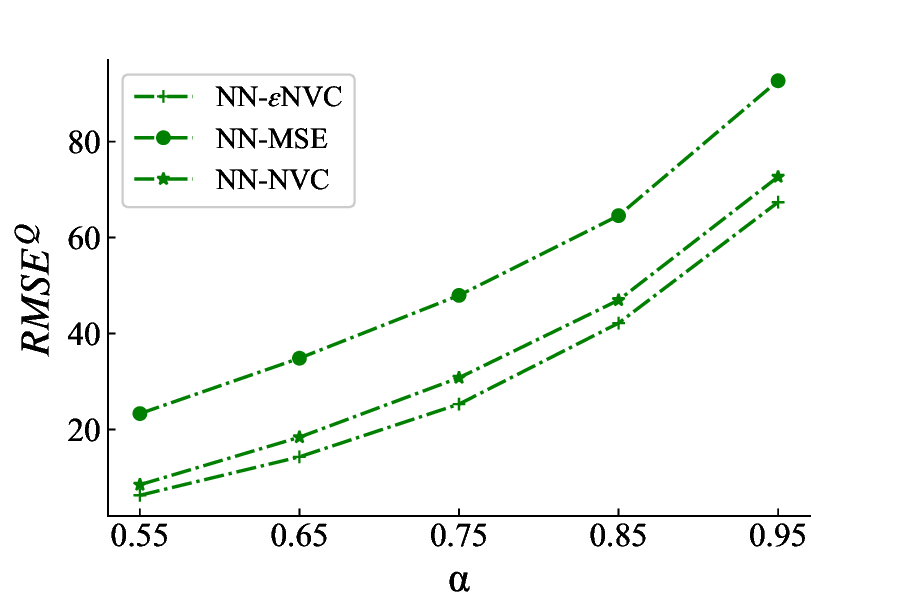}}
\caption{Out-of-Sample $\text{RMSE}^{\text{Q}}$}
\label{fig:testRMSEQ}
\end{figure}

To investigate the cost-saving mechanism of the algorithms using the $\varepsilon$-insensitive operational cost, we analyze the order quantities obtained by them.
We first compute the $RMSE^Q$, the root mean squared difference between the order quantity of the six algorithms and the optimal order quantity, as defined in Equation~~\eqref{eq:rmseq}.
The average $RMSE^Q$'s over 10 datasets of the six algorithms in different $\alpha$'s are shown in Figure \ref{fig:testRMSEQ}. 
The two algorithms with $\varepsilon$-insensitive newsvendor cost, namely, \LReNVCR ~and \NNeNVC, have the smallest $RMSE^Q$.
This means that the order quantities obtained by models with the $\varepsilon$-insensitive newsvendor cost loss are the closest to the optimal order quantities. 

We also conduct the paired samples Wilcoxon signed rank tests \citep{Wilcoxon1945IndividualCB} on $|\hat{y} - Q^*|$ between \LReNVCR ~(\NNeNVC) and \LRNVC ~(\NNNVC) for each dataset and each $\alpha$. 
The results are listed in Tables \ref{tab:wilcoxon-result-lr} and \ref{tab:wilcoxon-result-nn}.
The $|\hat{y}-Q^*|$'s of \LReNVCR ~and \NNeNVC ~are significantly smaller than those of \LRNVC ~and \NNNVC, respectively, with p-values less than 0.001. 
The Wilcoxon test results confirm the results in Figure~\ref{fig:testRMSEQ} statistically.

\begin{table}[!htbp]
\caption{Paired Samples Wilcoxon Signed Rank Tests on $|\hat{y}-Q^*|$ of \LReNVCR  ~and \LRNVC}
\label{tab:wilcoxon-result-lr}
\centering
\footnotesize
\begin{threeparttable}
\begin{tabular}{@{}llllll@{}}
\toprule
            & 0.55                       & 0.65                      & 0.75                       & 0.85                       & 0.95                       \\ \midrule
dataset 1  & $1.27 (1.03,1.59)^{***}$   & $1.19 (0.86, 1.47)^{***}$ & $1.37 (1.16, 18.83)^{***}$ & $1.10 (0.89, 30.27)^{***}$ & $3.68 (3.08, 42.59)^{***}$ \\
dataset 2  & $0.89 (-1.75, 1.58)^{***}$ & $0.65 (0.57, 0.85)^{***}$ & $0.74 (0.62, 20.22)^{***}$ & $1.00 (0.82, 31.33)^{***}$ & $0.76 (0.66, 9.86)^{***}$  \\
dataset 3  & $1.41 (-3.91, 1.62)$       & $1.00 (0.76, 1.26)^{***}$ & $0.97 (0.68, 14.31)^{***}$ & $0.82 (0.67, 27.01)^{***}$ & $0.54 (0.48, 4.08)^{***}$  \\
dataset 4  & $1.00 (-11.84, 1.17)$      & $1.29 (1.06, 1.42)^{***}$ & $1.24 (1.03, 20.52)^{***}$ & $1.01 (0.88, 35.17)^{***}$ & $0.70 (0.10, 29.34)^{***}$ \\
dataset 5  & $0.84 (-7.16, 1.88)$       & $1.28 (1.16, 1.38)^{***}$ & $0.85 (0.68, 19.11)^{***}$ & $0.96 (0.8, 31.24)^{***}$  & $0.94 (0.85, 9.79)^{***}$  \\
dataset 6  & $1.10 (-2.29, 1.34)^{***}$ & $1.30 (1.17, 1.46)^{***}$ & $0.73 (0.57, 18.01)^{***}$ & $4.72 (4.47, 29.09)^{***}$ & $0.84 (0.38, 28.34)^{***}$ \\
dataset 7  & $1.44 (-3.83, 1.75)$       & $0.68 (0.43, 0.93)^{***}$ & $1.03 (0.77, 17.34)^{***}$ & $0.50 (0.37, 29.73)^{***}$ & $0.32 (0.22, 8.78)^{***}$  \\
dataset 8  & $1.18 (-6.57, 3.65)$       & $0.94 (0.58, 1.24)^{***}$ & $1.76 (1.51, 17.32)^{***}$ & $1.48 (1.28, 31.59)^{***}$ & $1.04 (0.92, 11.08)^{***}$ \\
dataset 9  & $0.96 (-2.85, 2.21)^{***}$ & $0.96 (0.79, 1.16)^{***}$ & $0.97 (0.76, 15.24)^{***}$ & $0.61 (0.40, 30.88)^{***}$ & $0.58 (0.14, 34.35)^{***}$ \\
dataset 10 & $1.04 (-4.76, 3.87)^{**}$  & $0.96 (0.76, 1.16)^{***}$ & $1.26 (1.14, 17.64)^{***}$ & $1.14 (1, 32.38)^{***}$    & $0.91 (0.80, 12.43)^{***}$ \\ \bottomrule
\end{tabular}
\begin{tablenotes}
\footnotesize
\item[1] Data in table are ``median (1st quartile, 3rd quartile)" of the difference between $|\hat{y}-Q^*|$ of \LRNVC ~and that of \LReNVCR. 
\item[2] ***: $p\leq 0.001$; **: $p\leq 0.01$; *: $p\leq 0.05$.
\end{tablenotes}
\end{threeparttable}
\end{table}

\begin{table}[!htbp]
\caption{Paired Samples Wilcoxon Signed Rank Tests on $|\hat{y}-Q^*|$ of \NNeNVC  ~and \NNNVC}
\label{tab:wilcoxon-result-nn}
\centering
\footnotesize
\begin{threeparttable}
\begin{tabular}{@{}llllll@{}}
\toprule
            & 0.55                      & 0.65                      & 0.75                      & 0.85                      & 0.95                      \\ \midrule
dataset 1  & $2.11 (0.54,3.61)^{***}$  & $2.11 (0.54, 3.61)^{***}$ & $3.39 (2.41, 6.08)^{***}$ & $5.65 (4.5, 10.1)^{***}$  & $5.65 (4.5, 10.10)^{***}$ \\
dataset 2  & $2.78 (1.36, 4.75)^{***}$ & $2.78 (1.36, 4.75)^{***}$ & $6.82 (5.98, 8.16)^{***}$ & $1.31 (0.66, 3.15)^{***}$ & $1.31 (0.66, 3.15)^{***}$ \\
dataset 3  & $4.59 (3.64, 6.24)^{***}$ & $4.59 (3.64, 6.24)^{***}$ & $2.00 (1.19, 4.26)^{***}$ & $4.38 (3.28, 8.79)^{***}$ & $4.38 (3.28, 8.79)^{***}$ \\
dataset 4  & $4.33 (2.88, 6.25)^{***}$ & $4.33 (2.88, 6.25)^{***}$ & $3.01 (2.09, 6.05)^{***}$ & $3.03 (2.07, 6.81)^{***}$ & $3.03 (2.07, 6.81)^{***}$ \\
dataset 5  & $3.55 (2.10, 5.87)^{***}$ & $3.55 (2.10, 5.87)^{***}$ & $4.13 (3.17, 7.14)^{***}$ & $2.11 (1.29, 6.91)^{***}$ & $2.11 (1.29, 6.91)^{***}$ \\
dataset 6  & $3.16 (1.86, 4.71)^{***}$ & $3.16 (1.86, 4.71)^{***}$ & $2.92 (1.77, 4.79)^{***}$ & $1.00 (0.10, 3.73)^{***}$ & $1.00 (0.10, 3.73)^{***}$ \\
dataset 7  & $5.30 (4.15, 6.46)^{***}$ & $5.30 (4.15, 6.46)^{***}$ & $3.44 (2.58, 5.42)^{***}$ & $3.36 (2.47, 5.52)^{***}$ & $3.36 (2.47, 5.52)^{***}$ \\
dataset 8  & $6.12 (4.87, 7.24)^{***}$ & $6.12 (4.87, 7.24)^{***}$ & $2.01 (1.03, 4.48)^{***}$ & $3.09 (1.98, 5.51)^{***}$ & $3.09 (1.98, 5.51)^{***}$ \\
dataset 9  & $5.23 (3.74, 6.73)^{***}$ & $5.23 (3.74, 6.73)^{***}$ & $7.17 (6.19, 9)^{***}$    & $3.12 (2.37, 6.50)^{***}$ & $3.12 (2.37, 6.50)^{***}$ \\
dataset 10 & $1.95 (0.93, 3.66)^{***}$ & $1.95 (0.93, 3.66)^{***}$ & $6.98 (6.09, 8.72)^{***}$ & $4.55 (3.66, 7.25)^{***}$ & $4.55 (3.66, 7.25)^{***}$ \\ \bottomrule
\end{tabular}
\begin{tablenotes}
\footnotesize
\item[1] Data in table are ``median (1st quartile, 3rd quartile)" of the difference between $|\hat{y}-Q^*|$ of \NNNVC ~and that of \NNeNVC. 
\item[2] ***: $p\leq 0.001$; **: $p\leq 0.01$; *: $p\leq 0.05$.
\end{tablenotes}
\end{threeparttable}
\end{table}

\subsubsection{Service Level.}
We use the overbooking rate to represent the service level reached by the ML models, i.e., the sample proportion for which $\hat{y}$ is greater than demand in the test set.
We calculate the average service level in 10 groups of experiments, then the absolute difference (gap) between these values and the target service levels ($\alpha$), and finally the percentage improvement in the service level gap of \LReNVCR ~(\NNeNVC) over \LRMSE ~(\NNMSE) and over \LRNVC ~(\NNNVC).
The results are summarized in Table~\ref{tab:serviceLevel}.
\LReNVCR ~and \NNeNVC ~achieve closer service levels to the target service levels, with positive percentage improvements.
Besides, the improvements of \LReNVCR ~(\NNeNVC) over EAS methods are larger than those of IEO methods, consistent with the out-of-sample newsvendor cost performance.
Accordingly, we propose that in the case of censoring, the newsvendor can correct her order quantity decisions by setting a reasonable service level to lower the newsvendor cost.

\begin{table}[!htbp]
\centering
\footnotesize
\setlength{\tabcolsep}{1.6mm} 
\caption{Percentage Improvement of the Gap in Service Level}
\label{tab:serviceLevel}
\begin{tabular}{@{}ccccc@{}}
\toprule
& \textbf{$\frac{([\NNMSEa]-[\NNeNVCa])}{[\NNMSEa]}$} & \textbf{$\frac{([\NNNVCa]-[\NNeNVCa])}{[\NNNVCa]}$} & \textbf{$\frac{([\LRMSEa]-[\LReNVCRa])}{[\LRMSEa]}$} & \textbf{$\frac{([\LRNVCa]-[\LReNVCRa])}{[\LRNVCa]}$} \\ \midrule
$\alpha=0.55$    & 95.63\%   & 89.52\%   & 84.47\%   & 61.74\%           \\
$\alpha=0.65$    & 65.98\%   & 37.44\%   & 81.51\%   & 66.43\%           \\
$\alpha=0.75$    & 52.78\%   & 24.47\%   & 66.89\%   & 46.38\%           \\
$\alpha=0.85$    & 41.46\%   & 14.53\%   & 60.29\%   & 38.67\%           \\
$\alpha=0.95$    & 38.89\%   & 11.46\%   & 55.99\%   & 21.34\%           \\ \bottomrule
\end{tabular}
\end{table}

\section{Conclusion}
\label{sec:conclusion}

We propose an offline decision-learning framework that prescribes decisions based on past unobserved censored observations.
The framework processes the censored values of the key variable and learns decisions from the censored data and related features without making assumptions about the distribution of the underlying problem. 
We design $\varepsilon$-insensitive operational costs to guide the learning process that allows some amount of deviations from the unobserved censored observations of the key variables to the decision.
Theoretically, we show that the new learning algorithms are uniformly stable and provide their generalization bounds. 
We implement the new costs using LR models and NNs in a newsvendor problem.
Our proposed framework can handle censored key observations better. It achieves lower newsvendor costs and obtains closer to optimal order quantities. 
We also see from the results that the order quantities generated by our proposed algorithms correspond to lower newsvendor costs as the degree of censoring increases.

\bibliographystyle{informs2014}
\bibliography{loss-func-censor}

\newpage
\setcounter{page}{1} 
\begin{APPENDICES}

\section{Proof of Results in Section~\ref{sec:CustML}}
\setcounter{table}{0} 
\renewcommand{\thetable}{\thesection.\arabic{table}} 
\setcounter{figure}{0} 
\renewcommand{\thefigure}{\thesection.\arabic{figure}} 
\setcounter{equation}{0} 
\renewcommand{\theequation}{\thesection.\arabic{equation}} 

\label{app:ProofCustML}
\textit{Proof of Proposition~\ref{prop:LR-grad}}: 
Denote by $\eta$ the learning rate for each iteration in \textit{stochastic gradient descent} (SGD), which is a hyperparameter.
The \textit{linear regression} (LR) model is defined as
\begin{equation}
    y_i = \sum_{j=1}^p{x_{i}^{j}\cdot \theta^j}.
\end{equation}
The objective function for training the LR with the loss function \eqref{eq:le1e2} is given by 
\begin{equation}
\begin{split}
    &\min_{\boldsymbol{\theta}}{\frac{1}{n}\sum_{i=1}^n{\inNVLoss_i (\boldsymbol{\theta}|(\boldsymbol{x}_i,s_i))}} \\
    = &\min_{\boldsymbol{\theta}}{\frac{1}{n}\sum_{i=1}^n{\left[(1-\alpha)\cdot (\sum_{j=1}^p{x_{i}^{j}\cdot \theta^j} - s_i - \varepsilon_1)^+ + \alpha \cdot (s_i + \varepsilon_2 - \sum_{j=1}^p{x_{i}^{j}\cdot \theta^j})^+ \right]}}.
\label{eq:LRObjinmae.w}
\end{split}
\end{equation}
Taking the partial derivative of $\inNVLoss_i$ with respect to $\theta^j$, we have
\begin{equation}
\frac{\partial \inNVLoss_i}{\partial \theta^j} = 
\begin{cases}
(1-\alpha)x_{i}^{j}, &\sum_{j=1}^p{x_{i}^{j}\cdot \theta^j} > s_i+\varepsilon_1,\\
-\alpha x_{i}^{j}, &\sum_{j=1}^p{x_{i}^{j}\cdot \theta^j} < s_i+\varepsilon_2, \\
0, & \text{otherwise}.
\end{cases}
\end{equation}
Let $\theta^{j (k)}$ be the value of $\theta^j$ and $y_i^{(k)} = \sum_{j=1}^p{x_{i}^{j}}\cdot\theta^{j (k)}$ be the predicted value of $y_i$ in the $k$-th iteration of the mini-batch gradient descent method \citep{bottou2018optML}. Suppose the batch dataset we use in the $k$-th iteration is $S_{n_k}^{(k)}$  with batch size $n_k$. Then, we get the update rule for \LReNVC ~in the $(k+1)$-th iteration,
\begin{equation}
\begin{split}
\theta^{j (k+1)} &= \theta^{j (k)} - \frac{\eta}{n_{k+1}}\sum_{i\in S_{n_{k+1}}^{(k+1)}}{\frac{\partial \inNVLoss_i}{\partial \theta^j}}\\
&= \theta^{j(k)}-\frac{\eta}{n_{k+1}}\sum_{i\in S_{n_{k+1}}^{(k+1)}}{\mathbb{I}(y_i^{(k)}> s_i+\varepsilon_1)(1-\alpha)x_{i}^{j}}-\frac{\eta}{n_{k+1}}\sum_{i\in S_{n_{k+1}}^{(k+1)}}{\mathbb{I}(y_i^{(k)}< s_i+\varepsilon_2)\alpha x_{i}^{j}},
\end{split}
\end{equation}
where $\mathbb{I}(\cdot)$ is the indicator function. $\square$

\textit{Proof of Proposition~\ref{prop:DNNRule}}: Recall that the proposed loss function for \NNeNVC ~is $\inNVLoss$.
By the forward propagation algorithm, the input and the output for node $j$ in layer $\layerN$ are $\zeta_j^{\layerN} = \sum_{k\in\mathbb{N}^{l-1}}{\theta_{j,k}^\layerN a_k^{\layerN-1}} +b_j^{\layerN}$, and $a_j^{\layerN} = \sigma^l(\zeta_j^{\layerN})$, respectively. Then, $\frac{\partial{\zeta_j^{\layerN}}}{\partial{\theta_{j,k}^{\layerN}}} = a_k^{\layerN-1}$.
The partial gradient for loss function $\inNVLoss_i$ with respect to weight $\theta_{j,k}^\layerN$ and bias $b_j^\layerN$ can be obtained by the back-propagation algorithm \citep{rumelhart1986backpropagating}.
For the $i$-th observation, by the Chain Rule, 
\begin{equation}
\frac{\partial \inNVLoss_i}{\partial \theta_{j,k}^\layerN} =\frac{\partial \inNVLoss_i}{\partial \zeta_j^{\layerN}} \cdot \frac{\partial \zeta_j^{\layerN}}{\partial \theta_{j,k}^\layerN} = \frac{\partial \inNVLoss_i}{\partial \zeta_j^{\layerN}} \cdot a_k^{\layerN-1},
\end{equation}
and 
\begin{equation}
\frac{\partial \inNVLoss_i}{\partial b_{j}^\layerN} =\frac{\partial \inNVLoss_i}{\partial \zeta_j^{\layerN}} \cdot \frac{\partial \zeta_j^{\layerN}}{\partial b_{j}^\layerN} = \frac{\partial \inNVLoss_i}{\partial \zeta_j^{\layerN}}.
\end{equation}
Denoting by
\begin{equation}
\delta_j^{i,\layerN} = \frac{\partial \inNVLoss_i}{\partial \zeta_j^{i,\layerN}} =\frac{\partial \inNVLoss_i}{\partial a_j^{i,\layerN}}\frac{\partial a_j^{i,\layerN}}{\partial \zeta_j^{i,\layerN}} 
    = \frac{\partial \inNVLoss_i}{\partial a_j^{i,\layerN}} \sigma'(\zeta_j^{i,\layerN}),
\end{equation}
for nodes in layer $\nLayers$ with activation function $\sigma(\tau) = \tau$, we have
\begin{equation}
\delta_1^{i,\nLayers} = \frac{\partial \inNVLoss_i}{\partial \zeta_1^{i,\nLayers}} =\frac{\partial \inNVLoss_i}{\partial a_1^{i,\nLayers}}
    =
    \begin{cases}  
    \alpha, &y_i\leq s_i+\varepsilon_2, \\
    1-\alpha, &y_i\geq s_i+\varepsilon_1, \\
    0, &y_i\in(s_i+\varepsilon_2,s_i+\varepsilon_1).
    \end{cases}.
\end{equation}
Next we use mathematical induction to obtain $\delta_j^{i,\layerN}$ for any layer $\layerN$ between 2 and $\nLayers-1$ from layer $\nLayers$ to layer 1.
Suppose we have all $\delta_j^{i,\layerN}$ in layer $\layerN$, then in layer $\layerN-1$,
\begin{equation}
\begin{split}
\delta_k^{i,\layerN-1} &= \sum_{j\in\mathbb{N}^l}{\frac{\partial \zeta_j^{i,\layerN}}{\partial \zeta_k^{i,\layerN-1}}\cdot \frac{\partial \inNVLoss_i}{\partial \zeta_j^{i,\layerN}}} \\
& = \sum_{j\in\mathbb{N}^l}{\theta_{j,k}^\layerN \sigma'(\zeta_k^{\layerN-1}) \delta_j^{i,l}}\\
& = \sigma'(\zeta_k^{\layerN-1}) \cdot \sum_{j\in\mathbb{N}^l}{\theta_{j,k}^\layerN  \delta_j^{i,l}}.
\end{split}
\end{equation}
Therefore, we get the gradients of our proposed $\varepsilon$-insensitive loss function with respect to the weights of the network for the $i$-th observation,
\begin{equation}
 \frac{\partial \inNVLoss_i}{\partial \theta_{j,k}^\layerN} = a_k^{i,\layerN-1} \cdot \delta_j^{i,\layerN},
\end{equation}
and the gradients with respect to the bias,
\begin{equation}
\frac{\partial \inNVLoss_i}{\partial b_{j}^\layerN} = \delta_j^{i,\layerN}.
\end{equation}
Proposition~\ref{prop:DNNRule} is obtained. $\square$

\section{Proof of Results in Section~\ref{sec:prop}}
\setcounter{table}{0} 
\renewcommand{\thetable}{\thesection.\arabic{table}} 
\setcounter{figure}{0} 
\renewcommand{\thefigure}{\thesection.\arabic{figure}} 
\setcounter{equation}{0} 
\renewcommand{\theequation}{\thesection.\arabic{equation}} 

\label{app:ProofProp}

\textit{Proof of Lemma~\ref{lem:CostBound}}:
Clearly, $(\alpha \vee (1-\alpha))(\bar{D}+\varepsilon_2)$ is an upper bound on $|\inNVLoss|$ for all $y,s\in [0,\bar{D}]$. If $s=0$ and $y=\bar{D}$, $|\inNVLoss(s,y)| = (1-\alpha)(\bar{D}-\varepsilon_1)\leq(1-\alpha)(\bar{D}+\varepsilon_2)$, and conversely, if $y=0$ and $s=\bar{D}$,  $|\inNVLoss(s,y)| = \alpha(\bar{D}+\varepsilon_2)$. Hence, the upper bound is attained. $\square$\

\textit{Proof of Lemma~\ref{lem:lipschitz}}:
Intuitively, $|\inNVLoss(s,y_1)-\inNVLoss(s,y_2)|\leq (\alpha \vee (1-\alpha))|y_1-y_2|$, and thus $\inNVLoss(\cdot,\cdot)$ satisfies the Lipschitz condition.$\square$

Before the proof of Proposition~\ref{prop:LR-stable}, we restate the definition of uniform stability in \citet[Definition 6]{bousquet2002stability}. Denote by $A_S$ the output of the algorithm $\mathcal{A}$ with dataset $S$, and by $S^{\!\setminus\!i}$ a set of $S$ without the $i$-th observation $(\boldsymbol{x}_i;s_i)$.

\begin{definition} [Uniform Stability, {\citet[Definition 6]{bousquet2002stability}}]
\label{def:us}
An algorithm $\mathcal{A}$ has uniform stability $\xi$ with respect to the loss function $\mathcal{L}$ if the following holds 
\begin{equation}
\forall S \in \mathcal{X} \times \mathcal{D}, \ \forall i \in \{1,\cdots,n\}, \ ||\mathcal{L}(A_S,\cdot)-\mathcal{L}(A_{S^{\!\setminus\!i}},\cdot)||_\infty \leq \xi.
\end{equation}
\end{definition}

\textit{Proof Proposition~\ref{prop:LR-stable}}:
We prove the uniform stability of \LReNVC ~similar to \cite[Proposition EC.1]{ban2019big}. 
Note that $\hat{\mathcal{R}}^{LR-\varepsilon NV}(\boldsymbol{\theta},S_n)$ in Equation~\eqref{eq:LRObjInMae.w} is piecewise linear and convex with respect to $\boldsymbol{\theta}$. $\partial \hat{\mathcal{R}}^{LR-\varepsilon NV}(\boldsymbol{\theta}_n;S_n)$ (subgradients of $\hat{\mathcal{R}}^{LR-\varepsilon NV}(\cdot;S_n)$ at $\boldsymbol{\theta}_n$) is compact, and $0\in \partial \hat{\mathcal{R}}^{LR-\varepsilon NV}(\boldsymbol{\theta}_n;S_n)$ by the optimality of $\boldsymbol{\theta}_n$. 
Then, similar to \cite[Proposition EC.1]{ban2019big}, with the results in Lemma~\ref{lem:CostBound} and Lemma~\ref{lem:lipschitz}, we get
\begin{equation}
\xi_n = \frac{p}{n} \cdot \frac{(\alpha \vee (1-\alpha))^2}{(\alpha \wedge (1-\alpha))}\cdot (\bar{D}+\varepsilon_2).
\end{equation}
It is obvious that $\xi_n$ decreases as $\frac{1}{n}$. Therefore, we get the desired result by Definition 6 of uniform stability in \cite{bousquet2002stability}. $\square$

\textit{Proof of Theorem~\ref{theorem:LR-GenBound}}:
The result in Theorem~\ref{theorem:LR-GenBound} follows from \cite[Theorem 12]{bousquet2002stability}, Lemma~\ref{lem:CostBound} and Proposition~\ref{prop:LR-stable} in this paper.
Following \cite[Remark 13 for Theorem 12]{bousquet2002stability}, the bound given in Theorem~\ref{theorem:LR-GenBound} is tight, since the stability parameter $\xi_n$ for \LReNVC ~scales as $\frac{1}{n}$. 
$\square$

\textit{Proof Proposition~\ref{prop:LR-R-stable}}:
By the Lipschitz property of $\mathcal{L}^{\varepsilon NV}(s,\cdot)$,
\begin{align}
\sup_{s\in \mathcal{D}} |\mathcal{L}^{\varepsilon NV}(s,q_{\boldsymbol{\theta_1}}(\boldsymbol{x})) - \mathcal{L}^{\varepsilon NV}(s,q_{\boldsymbol{\theta_2}}(\boldsymbol{x}))| 
\leq (\alpha \vee (1-\alpha)) |q_{\boldsymbol{\theta_1}}(\boldsymbol{x})) - q_{\boldsymbol{\theta_2}}(\boldsymbol{x}))|, & \forall q_{\boldsymbol{\theta_1}}(\boldsymbol{x}),q_{\boldsymbol{\theta_2}}(\boldsymbol{x})\in \mathcal{Q}.
\end{align}
And $\mathcal{L}^{\varepsilon NV}(s,\cdot)$ is convex with respect to its second argument. 
Hence, $\mathcal{L}: \mathcal{X}\times \mathcal{D}\rightarrow \mathbb{R}$ is $(\alpha\vee(1-\alpha))$-admissible \citep[Definition 19]{bousquet2002stability}.  
$\mathbb{R}^p$ is a reproducing kernel Hilbert space where the kernel is the standard inner product. 
Thus, $\kappa^2 = p X_{\max}^2$ in our case. Hence, by \cite[Theorem 22]{bousquet2002stability}, the algorithm (\LReNVCR) has uniform stability with parameter $\xi_n^r$ as given.

\textit{Proof of Theorem~\ref{theorem:LR-R-GenBound}}:
The result follows from \cite[Theorem 12]{bousquet2002stability}, Lemma~\ref{lem:CostBound} and Proposition~\ref{prop:LR-R-stable}.

\textit{Proof of Proposition~\ref{prop:DNN-UASBound}}:
We prove the bound of the UAS of \NNeNVC~ trained by the fixed-permutation SGD similar to \citep[Theorem 3.2]{bassily2020stability}. 
Let $S\simeq S'$ be two datasets that differ only in a single data point, $f^1_S = f_{S'}^1$ be the same initial model, and the trajectories $(f^t_S)_{t\in[n]}, (f^t_{S'})_{t\in[n]}$ associated with the fixed permutation SGD method on datasets $S$ and $S'$, respectively. Since the datasets $S\simeq S'$ are random, we may assume without loss of generality that the fixed permutation $\pi$ is arbitrary. By \citep[Lemma 3.1]{bassily2020stability}, and $||\nabla C^{\varepsilon}(f^t;S)-\nabla C^{\varepsilon}(f^t;S')||\leq a_t$, with $a_t = 2(\alpha\vee(1-\alpha))\cdot \mathbb{I}{\{(t\mod{n})=i\}}$, we have
\begin{equation}
\begin{split}
||f^t_S - f^t_{S'}||&\leq 2 (\alpha\vee(1-\alpha))\sqrt{\sum_{j=1}^{t-1}\eta^2}+4(\alpha\vee(1-\alpha))\sum_{j=1}^{\lfloor(t-1)/n\rfloor}{\eta}\\
& \leq 2 (\alpha\vee(1-\alpha))\sqrt{\sum_{j=1}^{t-1}\eta^2}+\frac{4}{n}(\alpha\vee(1-\alpha))\sum_{j=1}^{t-1}{\eta}.
\end{split}
\end{equation}
  Since the bound holds for all iterates, and for \NNeNVC~ trained by $K$-pass SGD with the fixed permutation, the number of iterations is $nK$. Therefore,
\begin{equation}
\sup_{S\simeq S'}(f_S - f_{S'}) \leq 2 (\alpha \vee (1-\alpha))(\eta\sqrt{nK}+2K\eta).
\end{equation}
Here, $f_S\triangleq \mathcal{A}(S)$ .
$\square$

\textit{Proof of Theorem~\ref{theorem:dnnGBound}}:
With Proposition~\ref{prop:DNN-UASBound}, the generalization bound in Theorem~\ref{theorem:dnnGBound} is obtained by \citep[Theorem 1.1]{feldman2019high}.
$\square$

\section{Store Selection}
\label{app:storeSelect}
\setcounter{table}{0} 
\renewcommand{\thetable}{\thesection.\arabic{table}} 
\setcounter{figure}{0} 
\renewcommand{\thefigure}{\thesection.\arabic{figure}} 
\setcounter{equation}{0} 
\renewcommand{\theequation}{\thesection.\arabic{equation}} 
In this section, we document the analysis we conducted for selecting the appropriate store in our study.

The real-world dataset contains transactions of Corporaci\'{o}n Favorita's 53 stores.
In the original Kaggle dataset, there are 8 tables.
We use \texttt{train.csv} and \texttt{items.csv} only.
\texttt{train.csv} contains transaction records from January 1, 2013, to August 15, 2017. Each observation is an item's sales on a day in a store. We choose transactions in 2016 in our study.
\texttt{items.csv} contains information about an item, including item number, product category, product class, and whether it is perishable. We use it to identify perishable products.
To match the newsvendor setting, we separate transactions of perishable products.
In each store, there are 9 product categories' transaction records. 
We group the records by date, store, and product category.

We intend to locate one store whose transactions are best estimated by a linear regression model with a normally distributed error term. By doing so, we assume our synthesized data represents a store whose product demand approximately follows the demand generation process, i.e., $D = \boldsymbol{\beta}^\intercal \boldsymbol{x} + D^r$, where $\boldsymbol{\beta}$ are the coefficients to be estimated, $\boldsymbol{x}$ is the feature vector, and $D^r \sim \mathcal{N}(\mu,\sigma^2)$ is the random noise. We fit this demand generation model with a store's transaction records to obtain the estimation of $\boldsymbol{\beta}$ and $(\mu,\sigma)$, and then use the estimated model to generate demand. We choose the available feature variable, namely \textit{Day of Week, Month of Year}, and \textit{Product Category}.

The \textit{root mean square error} (RMSE) of each store's linear regression model is in column \textbf{RMSE} Table~\ref{tab:LRResults52Stores}.
The RMSE divided by the median of sales in the corresponding store is in column \textbf{RMSE over Median}.
The adjusted $R^2$ is in column \textbf{Adjust $R^2$}.
The model significance statistics $F$ of each linear regression model are in columns \textbf{F}, and all the corresponding $p$-values are close enough to zero, which indicates that our fitting linear regression models are all statistically significant.
We choose the store with the smallest RMSE and $\frac{\text{RMSE}}{\text{median of sales}}$, which is Store 10.
The descriptive statistics of Store 10's sales are summarized in Table~\ref{tab:EDAofStore10}.
\begin{table}[thbp]
\renewcommand\arraystretch {0.5}
\centering
\footnotesize
\caption{The Descriptive Statistics of Store 10's Sales}
\label{tab:EDAofStore10}
\begin{tabular}{cccccccc}
\toprule
\textbf{Count} & \textbf{Mean} & \textbf{Std} & \textbf{Min} & \textbf{25\%} & \textbf{50\%} & \textbf{75\%} & \textbf{Max}  \\ \midrule
3237 & 149.21 & 112.76 & 1.00 & 45.00 & 142.60 & 221.00 & 917.62 \\
\midrule
\end{tabular}
\end{table}

Lastly, we fit the linear regression model \eqref{eq:LRGenerate} with Store 10's transaction records, which results in Table~\ref{tab:LRstore10}.
For the distribution of the random noise, we estimate the mean and standard deviation using the residuals of Model ~\eqref{eq:LRGenerate} on Store 10, and we obtain a mean of 0 and a standard deviation of 46.57.  
The QQ plot of the residuals is shown in Figure~\ref{Fig:ErrorQQplot10}.

\begin{equation}
    \label{eq:LRGenerate}
    \begin{split}
    	y = &\beta_0 + \beta_1 \cdot Category_1 + \beta_2  \cdot Category_2 +\beta_3 \cdot Category_3 + \beta_4 \cdot Category_4 + \beta_5 \cdot Category_5+ \\
        &\beta_6 \cdot Category_6 + \beta_7 \cdot Category_7 + \beta_8 \cdot Category_8 + \beta_9 \cdot Tue + \beta_{10} \cdot Web + \beta_{11} \cdot Thur +\\
         &beta_{12} \cdot Fri + \beta_{13} \cdot Sat + \beta_{14} \cdot Sun +\beta_{15} \cdot Feb + \beta_{16} \cdot Mar +\beta_{17} \cdot Apr +\beta_{18} \cdot May + \\
         &\beta_{19} \cdot Jun + \beta_{20} \cdot Jul + \beta_{21} \cdot Aug + \beta_{22} \cdot Sept +\beta_{23} \cdot Oct +\beta_{24} \cdot Nov + \beta_{25} \cdot Dec.\\
    \end{split}
\end{equation}

\begin{table}[!htbp]
\renewcommand\arraystretch {0.5}
\centering
\footnotesize
\caption{The Linear Regression Result of Store 10}
\label{tab:LRstore10}
\begin{tabular}{cc|cc|cc}
\toprule
\begin{tabular}[c]{@{}c@{}}$\boldsymbol{X}$\\(\textit{Product Category})\end{tabular} & $\boldsymbol{\hat{\beta}}$ &\begin{tabular}[c]{@{}c@{}}$\boldsymbol{X}$\\(\textit{Day of Week})\end{tabular}  & $\boldsymbol{\hat{\beta}}$ & \begin{tabular}[c]{@{}c@{}}$\boldsymbol{X}$\\(\textit{Month of Year})\end{tabular}& $\boldsymbol{\hat{\beta}}$ \\ \midrule
1 & 192.23 & Tue. & -3.64 & Feb. & -3.46 \\ 
2 & 151.66 & Wed. & -25.41   &  Mar.  &  1.57 \\
3 & -57.3 & Thur. & -29.90 & Apr.  & 11.94  \\ 
4 & 51.56  & Fri. & -32.75  & May.  & 7.88 \\ 
5 & 55.42  & Sat. & 21.15   & Jun.  & -1.58  \\ 
6 & -76.14  & Sun. & 38.13  & Jul.  & -13.21   \\ 
7 & 130.65  &               &         & Aug.  & -11.9  \\
8 & -106.29  &               &         & Sept.  & 1.96 \\
              &          &               &         & Oct. & -1.67 \\
                &          &               &         & Nov. & -3.48  \\
 \textbf{Intercept}  & 113.40 &               &         & Dec. & 20.03 \\
\bottomrule
\end{tabular}
\end{table}

\begin{figure}[!htbp]  
    \centering
    \includegraphics[scale = 0.8]{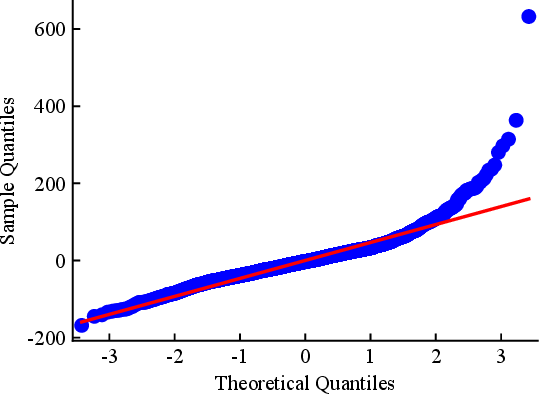}
    \caption{QQ Plot of Residual Error for Store 10 with Normal Distribution $\mathcal{N}(0,46.57^2)$}
    \label{Fig:ErrorQQplot10}
\end{figure}

\begin{table}[!htbp]
\caption{The Linear Regression Mode Results of 52 stores}
\label{tab:LRResults52Stores}
\renewcommand\arraystretch {0.5}
\centering
\footnotesize
\begin{tabular}{ccccccc}
\toprule
\textbf{No.} & \textbf{StoreID} & \textbf{RMSE} & \textbf{\begin{tabular}[c]{@{}c@{}}RMSE\\ over\\ Median\end{tabular}} & \textbf{$R^2$} & \textbf{\begin{tabular}[c]{@{}c@{}}Adjust\\ $R^2$\end{tabular}} & \textbf{F} \\ \midrule
1   &  10        &  46.5746       & 0.3266     &  0.8293     &  0.8280     &  624.1229        \\
2   &  14        &  54.6872       & 0.3275     &  0.8809     &  0.8800     &  954.0434        \\
3   &  11        &  162.8400      & 0.3346     &  0.8393     &  0.8381     &  678.7941        \\
4   &  13        &  48.3584       & 0.3838     &  0.7714     &  0.7696     &  426.2959        \\
5   &  22        &  46.6093       & 0.4270     &  0.8533     &  0.8522     &  739.5047        \\
6   &  9         &  193.7826      & 0.4394     &  0.8586     &  0.8576     &  789.4428        \\
7   &  19        &  62.2407       & 0.4480     &  0.8533     &  0.8522     &  752.1771        \\
8   &  23        &  119.2109      & 0.4610     &  0.9138     &  0.9131     &  1377.9255       \\
9   &  50        &  205.3945      & 0.4646     &  0.9213     &  0.9207     &  1521.6989       \\
10  &  15        &  66.8268       & 0.4673     &  0.8880     &  0.8871     &  1023.4136       \\
11  &  5         &  111.9503      & 0.4999     &  0.9361     &  0.9356     &  1903.2669       \\
12  &  46        &  395.1271      & 0.5131     &  0.8486     &  0.8474     &  728.4465        \\
13  &  48        &  328.5656      & 0.5231     &  0.8218     &  0.8204     &  599.5495        \\
14  &  16        &  69.1041       & 0.5618     &  0.8413     &  0.8400     &  661.6522        \\
15  &  6         &  260.9393      & 0.5737     &  0.9101     &  0.9094     &  1315.4281       \\
16  &  54        &  66.2729       & 0.5784     &  0.8840     &  0.8831     &  942.8589        \\
17  &  45        &  574.1935      & 0.5933     &  0.9208     &  0.9201     &  1510.4271       \\
18  &  32        &  52.3259       & 0.5946     &  0.6270     &  0.6240     &  211.3298        \\
19  &  34        &  110.4864      & 0.5988     &  0.8614     &  0.8603     &  797.2013        \\
20  &  2         &  239.4270      & 0.6028     &  0.9177     &  0.9171     &  1449.6547       \\
21  &  17        &  182.6757      & 0.6089     &  0.8344     &  0.8331     &  650.8533        \\
22  &  43        &  122.8362      & 0.6142     &  0.7598     &  0.7578     &  379.5114        \\
23  &  26        &  82.8409       & 0.6322     &  0.7878     &  0.7861     &  479.5208        \\
24  &  21        &  135.5839      & 0.6751     &  0.8662     &  0.8651     &  836.7606        \\
25  &  40        &  123.3867      & 0.6754     &  0.8763     &  0.8754     &  907.0674        \\
26  &  47        &  560.4408      & 0.6768     &  0.9156     &  0.9149     &  1410.2051       \\
27  &  30        &  79.6000       & 0.6922     &  0.7918     &  0.7902     &  489.9277        \\
28  &  4         &  239.8729      & 0.7276     &  0.9111     &  0.9104     &  1331.6203       \\
29  &  27        &  236.7949      & 0.7430     &  0.9220     &  0.9214     &  1536.3398       \\
30  &  28        &  228.6058      & 0.7545     &  0.8131     &  0.8117     &  565.1187        \\
31  &  35        &  89.0039       & 0.7731     &  0.5892     &  0.5857     &  169.2330        \\
32  &  3         &  608.6074      & 0.7767     &  0.9446     &  0.9442     &  2215.7184       \\
33  &  7         &  365.5852      & 0.7767     &  0.9541     &  0.9537     &  2701.3525       \\
34  &  33        &  175.4642      & 0.7890     &  0.8170     &  0.8155     &  559.4922        \\
35  &  20        &  252.0826      & 0.7929     &  0.6563     &  0.6537     &  248.2877        \\
36  &  49        &  588.5328      & 0.8108     &  0.9457     &  0.9452     &  2262.1900       \\
37  &  41        &  147.3007      & 0.8168     &  0.9380     &  0.9375     &  1966.8681       \\
38  &  24        &  331.8992      & 0.8950     &  0.9184     &  0.9177     &  1462.4586       \\
39  &  29        &  188.4952      & 0.9032     &  0.8116     &  0.8101     &  558.9625        \\
40  &  44        &  863.0062      & 0.9329     &  0.9250     &  0.9244     &  1602.7195       \\
41  &  18        &  198.6834      & 0.9428     &  0.8556     &  0.8542     &  609.6545        \\
42  &  8         &  433.3750      & 0.9547     &  0.9228     &  0.9222     &  1553.0564       \\
43  &  42        &  182.8367      & 0.9655     &  0.9451     &  0.9447     &  2237.9162       \\
44  &  51        &  528.4183      & 0.9893     &  0.9019     &  0.9011     &  1195.0267       \\
45  &  36        &  221.3055      & 1.0327     &  0.8054     &  0.8039     &  515.4907        \\
46  &  37        &  231.4815      & 1.1438     &  0.9352     &  0.9347     &  1876.5226       \\
47  &  53        &  277.5855      & 1.1871     &  0.8504     &  0.8492     &  734.1657        \\
48  &  12        &  206.6159      & 1.2744     &  0.2945     &  0.2891     &  53.8586         \\
49  &  1         &  263.5151      & 1.2986     &  0.8806     &  0.8797     &  959.1935        \\
50  &  38        &  219.5359      & 1.4162     &  0.9022     &  0.9014     &  1198.6081       \\
51  &  25        &  322.0701      & 2.1535     &  0.6835     &  0.6806     &  237.9330        \\
52  &  39        &  1601.9278     & 6.3781     &  0.0986     &  0.0916     &  14.2130        \\
\bottomrule
\end{tabular}
\end{table}

\section{Training Machine Learning Algorithms}
\label{app:training}
\setcounter{table}{0} 
\renewcommand{\thetable}{\thesection.\arabic{table}} 
\setcounter{figure}{0} 
\renewcommand{\thefigure}{\thesection.\arabic{figure}} 
\setcounter{equation}{0} 
\renewcommand{\theequation}{\thesection.\arabic{equation}} 
In this section, we introduce the training process. The general training process is summarized in the following three steps.

\begin{enumerate}[Step 1.]
    \item \textbf{Hyperparameters Optimization.} Under $\alpha = 0.55$ (and $\varepsilon_1 = \varepsilon_2 = 0$ for models with $\varepsilon$-insensitive operational cost functions), the hyperparameters of machine learning models are tuned using cross-validation method and \texttt{Hyperopt} framework \citep{bergstra2013hyperopt}. 
    We implement hyperparameter optimization in the training set.
    
    \item \textbf{Insensitive Parameters Optimization.} For ML models with the standard newsvendor loss function and MSE loss function, skip this step and go to the next step. For ML models with $\varepsilon$-insensitive operational cost functions, the insensitive parameters (i.e., $\varepsilon_1, \varepsilon_2$) are tuned for each $\alpha$ using cross-validation method and \texttt{Hyperopt} framework \citep{bergstra2013hyperopt}. Also, we use the training set in this optimization phase.
    
    \item \textbf{Model Fitting and Evaluation.} For each $\alpha$, setting the hyperparameters and insensitive parameters as the optimized results obtained from Steps 1 and 2, we fit the ML models in the training set with censored data (i.e. sales data) and perform evaluations on the test set with uncensored data (i.e. demand data). We use the uncensored data in the test set to compute the out-of-sample newsvendor cost and $RMSE^{Q}$. 

\end{enumerate}

Next, we detail the training process for the six models.
For LR-based models, there is no hyperparameter in \LRMSE, and we skip the above Steps 1 and 2 and directly go to Step 3. For \LRNVC ~and \LReNVCR, we use the mini-batch GD algorithm to fit them. Since there is no existing python package, we code the training processes for \LRNVC ~and \LReNVCR ~by ourselves, and the hyperparameters in the mini-batch GD algorithm for \LRNVC ~and \LReNVCR ~are listed in Table~\ref{tab:LRHyperparam}.

\begin{table}[!htbp]
\footnotesize
\renewcommand\arraystretch {0.5}
\centering
\caption{Hyperparameters in mini-batch GD Algorithm for \LRNVC ~and \LReNVCR}
\label{tab:LRHyperparam}
\begin{tabular}{@{}lll@{}}
\toprule
\textbf{Notation} & \textbf{Meaning}                                        & \textbf{Value} \\ \midrule
eta               & Learning rate                                           & Tuned in training   \\
beta              & Coefficient of L2 regularization                        & Tuned in training    \\
batch\_size       & Number of samples used for each weight updating in SGD  & Tuned in training    \\
epoch             & Maximum number of iterations                            & $500$          \\
val\_size         & Ratio of validation set                                 & $0.25$         \\
shuffle  & \makecell*[l]{Whether to shuffle the dataset before split into train set \\ and validation set}                                    & True \\
random\_state     & Random seed                                             & $123$          \\
earlyStopping     & Whether to use early stopping                           & True           \\
tolerance         & Accuracy of judging whether the loss is reduced         & $0.0000001$    \\
patience & \makecell*[l]{If the loss of continuous ``patience" epoch was not \\reduced and ``earlyStopping" equals True, stop the \\ iterations} & $30$ \\
baseline          & If the loss is less than baseline, stop the iterations. & $0.000001$     \\
alpha             & Coefficient in the loss function                        & $0.5$          \\ \bottomrule
\end{tabular}
\end{table}

For NN-based models, we use \texttt{Keras} library (\href{https://keras.io}{https://keras.io}) in Python. 
We can directly use \texttt{Keras} to train \NNMSE ~without customization. For \NNNVC ~and \NNeNVC, we customize the proposed loss functions according to API provided by \texttt{Keras}. For each NN-based model, we use a network with two hidden layers. The customized hyperparameters in NN-based models are listed in Table~\ref{tab:DNNHyperparam}. The rest of the hyperparameters except for ``loss" and ``metric" in NN-based models use the default values in \texttt{Keras} package.

\begin{table}[!htbp]
\footnotesize
\renewcommand\arraystretch {0.5}
\centering
\caption{Custom Hyperparameters in NN for \NNMSE, \NNNVC, and \NNeNVC}
\label{tab:DNNHyperparam}
\begin{tabular}{@{}lll@{}}
\toprule
\textbf{Notation}    & \textbf{Meaning}                                       & \textbf{Value}       \\ \midrule
units1      & Number of nodes in hidden layer 1             & Tuned in training \\
units2      & Number of nodes in hidden layer 2             & Tuned in training \\
batch\_size & Number of samples per gradient update         & Tuned in training \\
nb\_epochs  & Maximum number of iterations                  & 500         \\
optimizer   & Optimizer used in gradient update             &``adam"      \\
activation  & activation function in nodes of hidden layers & ``sigmoid"   \\ \bottomrule
\end{tabular}
\end{table}

To specify the process of hyperparameter optimization, we take \NNeNVC ~as an example. We denote training set by $Dtrain$, kfolds in cross-validation by $V$, searching space for hyperparameters of ML models by $\Pi$, and the optimal hyperparameters of ML models by $\pi^*$. The customized hyperparameter optimization algorithm for \NNeNVC ~is detailed in Algorithm~\ref{alg:DNNinNVe}.

\clearpage
\begin{minipage}{15cm} 
\centering
\begin{algorithm}[H] 
\caption{\small{Customized Hyperparameter Optimization Algorithm for \NNeNVC}} 
\label{alg:DNNinNVe}
\footnotesize
\begin{algorithmic} [1]
\Require
    $Dtrain = (Xtrain,ytrain)$; 
    $V$;
    $\Pi$.
    
\Ensure
    $\pi^*$.
\Function{HyperoptHyperparameterOptimization}{$Dtrain,\Pi,V,\alpha,\varepsilon_1,\varepsilon_2$}
    \State Define the hyperparameters searching space $\Pi$
    \Function{ObjectiveFunctionOfHyperopt}{$\pi_p$}
        \For{$j=1$ to $V$}
                        \State Split $Dtrain \rightarrow 75\% Dtrain_{j} + 25\% Dtest_{j}$
                        \State Train the \NNeNVC ~machine with hyperparameters $\pi_p$ on $Dtrain_{j}$
                        \State Obtain the predicted values $\hat{y}_{j}^{test}$ on $Dtest_{j}$
                        \State Calculate the average $\varepsilon$-insensitive newsvendor cost $inNVCost_{j,p}$ on $Dtest_{j}$
                    \EndFor
        \State \Return{$inNVcost_{p} = \frac{1}{V}\sum_{j=1}^{V}inNVcost_{j,p}$}
    \EndFunction
    \State Call \texttt{fmin} function from \texttt{Hyperopt} package to select the optimal $\pi^* \in \Pi$ on $Dtrain$
    \Statex \qquad according to \Call{ObjectiveFunctionOfHyperopt}{$\pi_p$}
    \State \Return{$\pi^*$}
\EndFunction

\State $\alpha \leftarrow 0.55, \varepsilon_1 \leftarrow 0.0, \varepsilon_2 \leftarrow 0.0$.
\State $\pi^* \leftarrow$ \Call{HyperoptHyperparameterOptimization} {$Dtrain,\Pi,V,\alpha,\varepsilon_1,\varepsilon_2$}
\end{algorithmic}
\end{algorithm}
\end{minipage}

\vspace{2em} 

The searching space for ML models' hyperparameters and the results of Step 1 are summarized in Table~\ref{tab:searchingSpace}, in which $U(a,b)$ denotes a uniform distribution from $a$ to $b$.

\begin{table}[!htbp]
\centering
\renewcommand\arraystretch {0.5}
\footnotesize
\caption{Searching Spaces and Results for Hyperparameters of Step 1}
\label{tab:searchingSpace}
\begin{tabular}{@{}lllc@{}}
\toprule
\textbf{Model}                                        & \textbf{Hyperparameters} & \textbf{Searching space} & \multicolumn{1}{l}{\textbf{Result of CV}} \\ \midrule
\multirow{3}{*}{\LRNVC, \LReNVCR}                      & eta                       & $U(0.005,0.025)$         & 0.0007                                           \\
                                                      & beta                      & $U(0.0, 0.001)$          & 0.0060                                           \\
                                                      & batch\_size               & $\{64,65,… ,256\}$       & 98                                               \\ \midrule
\multicolumn{1}{l}{\multirow{3}{*}{\NNMSE}}          & units1                    & $\{4,5,…,10\}$           & 9                                                \\
\multicolumn{1}{l}{}                                 & units2                    & $\{3,4,…,8\}$            & 7                                                \\
\multicolumn{1}{l}{}                                 & batch\_size               & $\{64,65,…,256\}$        & 93                                               \\ \midrule
\multicolumn{1}{l}{\multirow{3}{*}{\NNNVC, \NNeNVC}} & units1                    & $\{4,5,…,10\}$           & 9                                                \\
\multicolumn{1}{l}{}                                 & units2                    & $\{3,4,…,8\}$            & 5                                                \\
\multicolumn{1}{l}{}                                 & batch\_size               & $\{64,65,…,256\}$        & 79                                               \\ \bottomrule
\end{tabular}
\end{table}

The process in Step 2 for tuning the insensitive parameters, i.e. $\varepsilon_1$ and $\varepsilon_2$, is similar to Algorithm~\ref{alg:DNNinNVe}, except that, in Step 2, we should set the searching space for insensitive parameters, and seize the optimal insensitive parameters for each $\alpha$.
In the ML models with loss function $\inNVLoss$, the searching spaces for $\varepsilon_1$ and $\varepsilon_2$ are set as $U(0,0.20)$ and $U(0,0.15)$, respectively. The search results for $\varepsilon_1$ and $\varepsilon_2$ are shown in Table~\ref{tab:epsilonSearchResults}.

\begin{table}[!htbp]
\footnotesize
\renewcommand\arraystretch {0.5}
\centering
\caption{Searching Results for $\varepsilon_1$ and $\varepsilon_2$ under Different $\alpha$'s}
\label{tab:epsilonSearchResults}
\begin{tabular}{@{}ccccccccccc@{}}
\toprule
         & \multicolumn{5}{c}{$\varepsilon_1$}        & \multicolumn{5}{c}{$\varepsilon_2$}        \\ \cmidrule(lr){2-6}  \cmidrule(lr){7-11}
\textbf{$\alpha$} & 0.55   & 0.65   & 0.75   & 0.85   & 0.95   & 0.55   & 0.65   & 0.75   & 0.85   & 0.95   \\ \midrule
\textbf{\LReNVC}  & 0.1976 & 0.1976 & 0.1976 & 0.1976 & 0.1827 & 0.0022 & 0.0022 & 0.0022 & 0.0022 & 0.0083 \\
\textbf{\NNeNVC}  & 0.1976 & 0.1976 & 0.1976 & 0.1976 & 0.1976 & 0.0022 & 0.0022 & 0.0022 & 0.0022 & 0.0022 \\ \bottomrule
\end{tabular}
\end{table}


\end{APPENDICES}
\end{document}